\documentclass[11pt]{article}
\usepackage[preprint]{acl}
\usepackage{times}
\usepackage{latexsym}
\usepackage[T1]{fontenc}
\usepackage[utf8]{inputenc}
\usepackage{microtype}
\usepackage{inconsolata}
\usepackage{graphicx}
\usepackage{xspace}
\usepackage{pifont}
\usepackage{amsmath}
\usepackage{amssymb}
\usepackage{array}
\usepackage{booktabs}
\usepackage{multirow}
\usepackage{colortbl}
\usepackage{tcolorbox}
\tcbuselibrary{listings}

\definecolor{algpurple}{HTML}{7A3FC8}
\definecolor{oursrowbg}{gray}{0.94}

\newenvironment{questionbox}{%
  \begin{quote}\itshape
}{%
  \end{quote}
}

\graphicspath{{figures/}}

\newcommand{\method}{\textsc{EvoSkill-S1}\xspace}
\newcommand{\opfusion}{Complementary Skill Bank Fusion}
\newcommand{\grpo}{\textsc{GRPO}\xspace}
\newcommand{\tablehead}[1]{\shortstack[c]{#1}}
\newcommand{\ourscell}[1]{\cellcolor{oursrowbg}#1}
\newcounter{mainalg}
\newenvironment{mainalgorithm}[1]{%
  \begin{center}
  \refstepcounter{mainalg}%
  \setlength{\tabcolsep}{0pt}
  \begin{tabular}{@{}r@{\hspace{0.35em}}p{0.82\columnwidth}@{}}
    \toprule
    \multicolumn{2}{@{}p{\columnwidth}@{}}{\textbf{Algorithm \themainalg: #1}} \\
    \midrule
}{%
    \bottomrule
  \end{tabular}
  \end{center}
}

\title{\method: Evolving Reusable Skills with Genetic-Inspired Operators}

\author{
\textbf{Yujun Wang\textsuperscript{1,2,3,*}},
\textbf{Jiqiang Li\textsuperscript{4,*}},
\textbf{Yinuo Lu\textsuperscript{5}},
\textbf{Jinhe Bi\textsuperscript{1,2}},
\textbf{Aniri\textsuperscript{1}} \\
\textbf{Wenxuan Ye\textsuperscript{3,6}},
\textbf{Boliang Liu\textsuperscript{3,7}},
\textbf{Sikuan Yan\textsuperscript{1,2,3}},
\textbf{Shuning Wang\textsuperscript{4}} \\
\textbf{Xuebing Zhou\textsuperscript{3}},
\textbf{S{\"o}ren Pirk\textsuperscript{8}},
\textbf{Hinrich Sch{\"u}tze\textsuperscript{1,2}},
\textbf{Yunpu Ma\textsuperscript{1,2}} \\[0.45em]
\normalfont
\textsuperscript{1}LMU Munich;
\textsuperscript{2}Munich Center for Machine Learning (MCML); \\
\textsuperscript{3}Huawei Heisenberg Research Center;
\textsuperscript{4}Zhejiang University; \\
\textsuperscript{5}University of Hong Kong; \\
\textsuperscript{6}Technical University of Munich (TUM);
\textsuperscript{7}TU Berlin;
\textsuperscript{8}Kiel University. \\
\textsuperscript{*}Equal contribution.
}

\definecolor{promptolive}{HTML}{536347}
\definecolor{promptcream}{HTML}{FDF8EA}

\newtcblisting{promptbox}[1][]{%
  colback=promptcream,%
  colframe=promptolive,%
  colbacktitle=promptolive,%
  coltitle=white,%
  fonttitle=\small\bfseries,%
  boxrule=0.7pt,%
  arc=2.5mm,%
  left=6pt,%
  right=6pt,%
  top=6pt,%
  bottom=6pt,%
  listing only,%
  listing options={%
    basicstyle=\ttfamily\footnotesize,%
    breaklines=true,%
    columns=fullflexible,%
    keepspaces=true%
  },%
  title=#1%
}

\newcounter{appendixalg}
\newenvironment{appendixalgorithm}[1]{%
  \refstepcounter{appendixalg}%
  \begin{center}
  \setlength{\tabcolsep}{0pt}%
  \begin{tabular}{@{}r@{\hspace{0.35em}}p{0.80\columnwidth}@{}}
    \toprule
    \multicolumn{2}{@{}p{\columnwidth}@{}}{%
      \textbf{Algorithm \theappendixalg: #1}} \\
    \midrule
}{%
    \bottomrule
  \end{tabular}
  \end{center}
}

\begin{document}
\maketitle
\begingroup
\renewcommand{\thefootnote}{}
\footnotetext{\raggedright\textbf{Correspondence:} \texttt{yujun\_wang\_cn@hotmail.com},
  \texttt{cognitive.yunpu@gmail.com}.}
\endgroup


\begin{abstract}
Reusable skills allow language-model agents to accumulate and reuse problem-solving experience. However, most existing methods improve a skill bank only indirectly through rollout-driven accumulation and heuristic reflection, often resulting in shallow and weakly controlled updates. We present \method, an RL-driven framework that instead treats reusable skills themselves as the direct objects of evolution. Given hard-negative failures, \method selects parent skills and transforms them into candidate child skills through five genetic-inspired operators: revision,
cross-revision,
mutation,
dual-reflection, and 
crossover. Downstream task rewards train the skill evolver to generate more
effective child skills, while evaluator-guided write-back selectively
retains beneficial updates. \opfusion{} consolidates operator-specific banks to exploit their distinct recovery patterns, while a Procedural Skill Interface supports transfer to unseen problems. 
Across all reported math and code benchmarks, \method improves the
corresponding initialized skill banks under both Qwen and GPT-family
solvers, with relative gains of up to 12.39\%.
Compared with the strongest external baseline without direct skill
evolution, the largest relative improvement reaches 20.39\%.
\end{abstract}

\begin{figure}[!t]
  \centering
  \includegraphics[width=\linewidth]{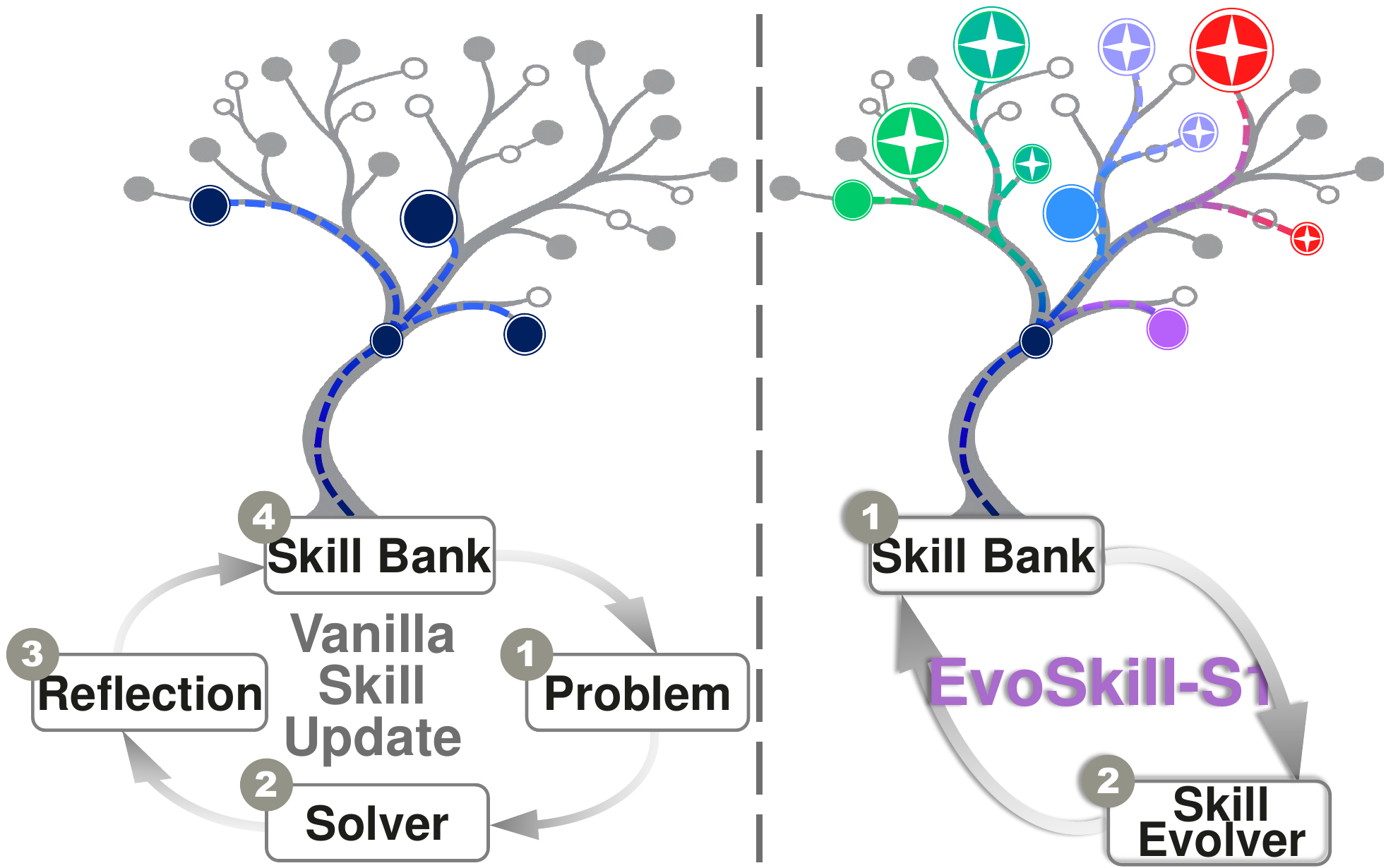}
  \caption{Heuristic skill-bank updates versus direct skill-level evolution in \method, where an RL-trained evolver transforms existing skills through genetic-inspired operators.}
  \label{fig:overview}
\end{figure}

\section{Introduction}

Large language models (LLMs) have become increasingly capable of complex reasoning, including mathematical problem solving and code generation. Nevertheless, many challenging problems still benefit from reusable problem-solving experience rather than isolated one-shot generation
\citep{brown2020language,wei2022chain,kojima2022large,wang2023selfconsistency,cobbe2021training,hendrycks2021math,chen2021evaluating,li2022competition}.
A natural way to provide such experience is through a skill bank: a collection of compact abstractions, strategies, and cautions that can be retrieved for related problems, echoing broader work on language-agent memory, tool use, and experience reuse
\citep{yao2023react,schick2023toolformer,park2023generative,wang2023voyager,packer2023memgpt,NEURIPS2023_1b44b878,zhao2024expel}.

Many existing approaches improve such banks only indirectly. As new rollouts arrive, they append new skills or summarize and reflect on additional experience
\citep{wang2023voyager,NEURIPS2023_1b44b878,zhao2024expel}.
This rollout-driven maintenance supports continual accumulation, but often leaves existing skills largely unchanged or only shallowly edited through weakly controlled heuristics. Consequently, a skill bank may continue to grow without systematically repairing brittle strategies, exploring alternative solution routes, or combining complementary partial skills.

We instead ask whether reusable skills themselves can become the direct objects of evolution.
In math and code, unresolved rollout failures provide concrete diagnostic signals: they expose where retrieved skills remain incomplete, brittle, or poorly matched to the target problem
\citep{cobbe2021training,lightman2023lets,chen2021evaluating,austin2021program,hendrycks2021apps,jain2024livecodebench}.
Rather than merely collecting more experience from these failures, they can be used to directly transform existing skills into stronger descendants.

Direct skill-level evolution, however, is non-trivial.
First, different failure modes require different transformations: some skills need conservative repair, while others require cross-problem transfer, an alternative solution route, stronger guardrails, or composition with another skill.
Second, a single hard negative can yield multiple plausible child
skills, requiring a principled mechanism for comparing candidates and
assigning credit to useful transformations. Third, evolved skills may
retain artifacts specific to their source problems, limiting transfer
to unseen problems. Finally, different operators may recover
complementary subsets of failures, requiring a mechanism to consolidate
their distinct strengths.

These challenges lead to the following question:

\begin{questionbox}
Can reusable skills themselves be directly and iteratively evolved
into stronger descendants while remaining efficient enough for
skill-bank improvement?
\end{questionbox}

In response, we propose \method, an RL-driven framework for direct skill-level evolution.
Without updating the base solver, \method uses hard negatives as evolution targets, selects parent and support skills from the current bank, and transforms them into candidate child skills through five genetic-inspired operators: revision, cross-revision, mutation, dual-reflection, and crossover.
Using \grpo-style group-relative feedback, the trainable skill evolver learns to favor child skills that improve downstream task performance, while evaluator-guided write-back retains beneficial updates and discards ineffective or redundant candidates.
Because different operators exhibit complementary repair patterns,
Complementary Skill Bank Fusion consolidates their operator-specific
banks under the same overall skill-bank budget.
A Procedural Skill Interface further supports transfer of evolved skills to unseen problems.

Experiments across math and code benchmarks show that \method
consistently improves the corresponding initialized skill banks under
both Qwen and GPT-family solvers, achieving relative gains of up to
12.39\%.
Compared with the strongest external baseline without direct skill
evolution, the largest relative improvement reaches 20.39\%.

This paper makes three main contributions:
\begin{itemize}
    \item \textbf{RL-guided direct skill evolution.}
    We formulate skill-bank improvement as an explicit optimization
    problem over reusable skills themselves and train a dedicated skill
    evolver with downstream task rewards to transform existing parent
    skills into stronger child skills, while keeping the base solver fixed.
    
    \item \textbf{Complementary genetic-inspired operators.}
    We introduce five genetic-inspired operators that apply distinct
    transformations to reusable skills. Their complementary recovery
    patterns address heterogeneous failure modes and motivate the fusion
    of operator-specific skill banks.
      
  \item \textbf{Complementary Skill Bank Fusion and empirical validation.}
  We consolidate operator-specific banks under a fixed overall budget,
  support the use of evolved skills on unseen problems through a
  Procedural Skill Interface, and demonstrate consistent improvements
  across math and code benchmarks.
\end{itemize}


\begin{figure*}[t]
  \centering
  \includegraphics[width=\linewidth]{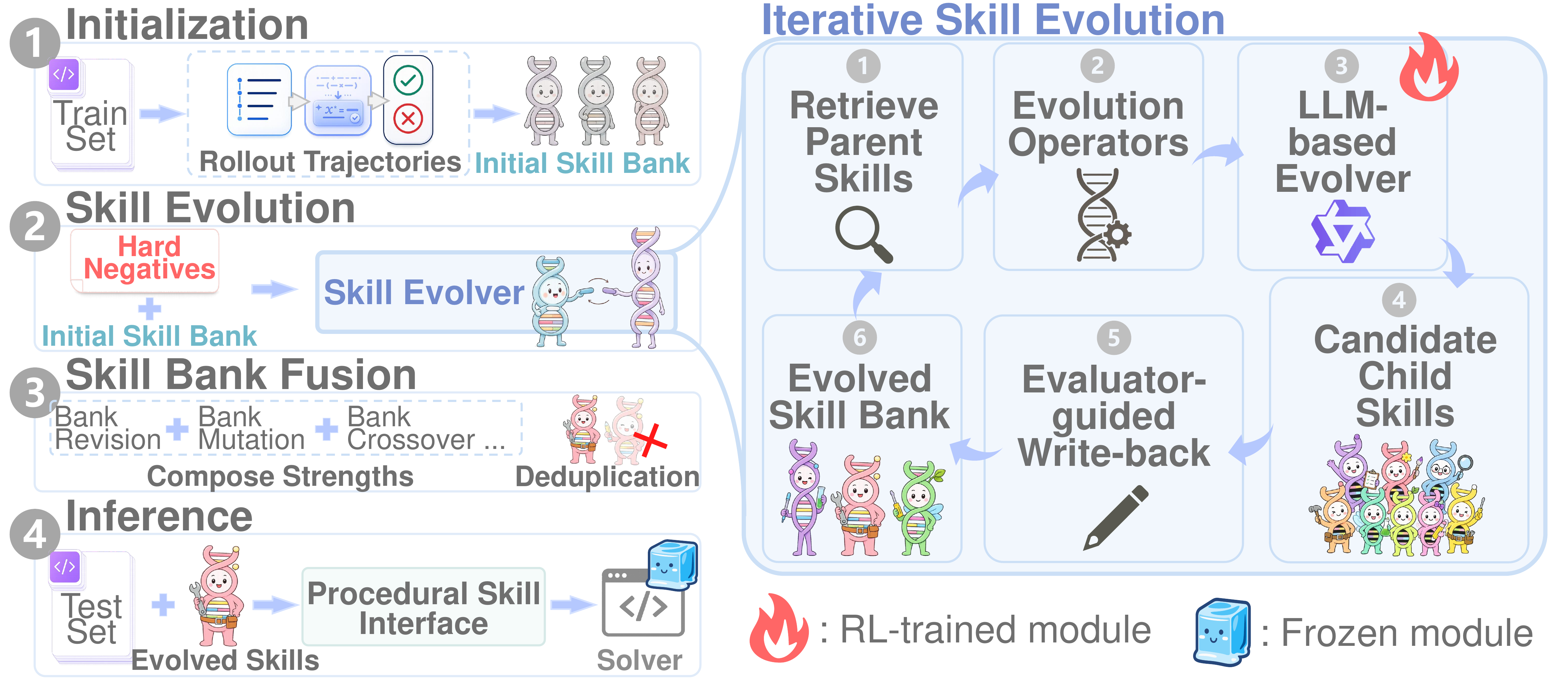}
  \caption{Overview of the \method pipeline. An initial skill bank is evolved through fixed-operator runs, where an RL-trained evolver generates candidate child skills and evaluator-guided write-back updates the bank. Operator-specific banks are then fused, and the resulting evolved
skills are accessed through the Procedural Skill Interface at
inference time.}
  \label{fig:workflow}
\end{figure*}

\section{Related Work}

\paragraph{LLM agents, memory, and retrieval.}
LLM agents often combine reasoning with actions, tools, environment
feedback, and persistent memory.
ReAct interleaves reasoning and acting
\citep{yao2023react}, Toolformer learns to invoke external tools
\citep{schick2023toolformer}, and Generative Agents maintain memory
streams for interactive simulation
\citep{park2023generative}.
Voyager further stores executable skills in an expanding library
\citep{wang2023voyager}, while MemGPT manages context outside model
parameters for later reuse \citep{packer2023memgpt}.
Retrieval-augmented methods ground generation in external evidence
\citep{lewis2020rag,karpukhin2020dense,guu2020realm,
borgeaud2022retro,gao2023rag}, and Self-RAG jointly controls retrieval
and self-critique \citep{asai2024selfrag}.
Together, these works establish external memory and retrieval as
effective mechanisms for accumulating reusable knowledge.

\paragraph{Experience learning and procedural skills.}
Reflexion stores verbal feedback from prior failures
\citep{NEURIPS2023_1b44b878}, while ExpeL extracts reusable insights from
training experience \citep{zhao2024expel}.
ReasoningBank distills generalizable reasoning strategies from
successful and failed trajectories and combines memory retrieval with
test-time scaling \citep{ouyang2026reasoningbank}.
Skill-Pro represents procedural skills through activation, execution,
and termination conditions, and uses semantic-gradient proposals with
a PPO-style gate to maintain the skill memory
\citep{mi2026skillpro}.
These approaches primarily derive, retrieve, or refine reusable
knowledge from episodic experience.

\paragraph{Skill and trajectory evolution.}
Recent work increasingly treats trajectories or skills as explicit
optimization objects.
SE-Agent evolves interaction trajectories through revision,
recombination, and refinement \citep{guo2025seagent}.
SkillRL constructs a hierarchical skill library and co-evolves it with
the solving policy through reinforcement learning
\citep{xia2026skillrl}.
Skill-R1 trains a lightweight skill generator for recurrent
instance-level skill optimization using group-relative feedback across
rollouts and successive revisions \citep{vishe2026skillr1}.
MetaSkill-Evolve operates at two timescales, improving both task skills
and the meta-skill that controls the improvement pipeline
\citep{wang2026metaskillevolve}.
In contrast, \method keeps the base solver fixed and trains a dedicated
evolver whose optimization targets are the reusable skills in a
persistent bank, rather than the solving policy or the trajectories
from which skills are derived.

\section{Problem Setup and Motivation}
\label{sec:setup-motivation}

\subsection{Problem Setup}

\paragraph{Task setting.}
Let $\mathcal{D}=\{(x_i,y_i)\}_{i=1}^{N}$ denote a collection of
tasks, where $x_i$ is the problem input and $y_i$ is the reference
answer or executable specification.
At evolution step $t$, a retriever scores skills with $\sigma(x,s)$ and selects the top-$k$ retrieved skill set
\begin{equation}
  \mathcal{R}_t(x)=\operatorname*{arg\,topk}_{s\in\mathcal{S}_t} \sigma(x,s).
\end{equation}
The solver then samples an answer conditioned on these retrieved skills, and the evaluator maps the answer to a task reward:
\begin{equation}
\begin{aligned}
  a_t(x) &\sim \pi_{\theta}(\cdot\mid x,\mathcal{R}_t(x)),\\
  r_t(x) &= E(x,y,a_t(x)).
\end{aligned}
\end{equation}
Here $E$ denotes exact match, unit-test pass rate, or another benchmark-specific evaluator.

\paragraph{Skill bank initialization.}
The initial skill bank is constructed from rollout data.
We collect base-solver rollouts
\begin{equation}
\begin{aligned}
  \mathcal{T}_0
  &= \{(x_i,a_0(x_i),r_0(x_i))\}_{i=1}^{M},\\
  \mathcal{S}_0
  &= \mathrm{Curate}_{\phi_0}
  \bigl(\mathcal{T}_0\bigr),
\end{aligned}
\end{equation}
where $\mathrm{Curate}_{\phi_0}$ denotes an LLM-based skill extraction
procedure performed by the initial skill evolver $q_{\phi_0}$ before
reinforcement-learning-based evolution.
The evolver summarizes reusable problem-solving guidance from the
rollout trajectories to form the initial skill bank.
For math, we run the base solver on roughly 17k DAPO problems
\citep{yu2025dapo} and extract the initial math skills from the
resulting trajectories.
For code, we apply the same pipeline to the APPS training set
\citep{hendrycks2021apps}.
The resulting banks serve as the starting points for subsequent skill
evolution.

\subsection{Why Directly Evolve Initialized Skills?}

\begin{figure}[t]
  \centering
  \includegraphics[width=\linewidth]{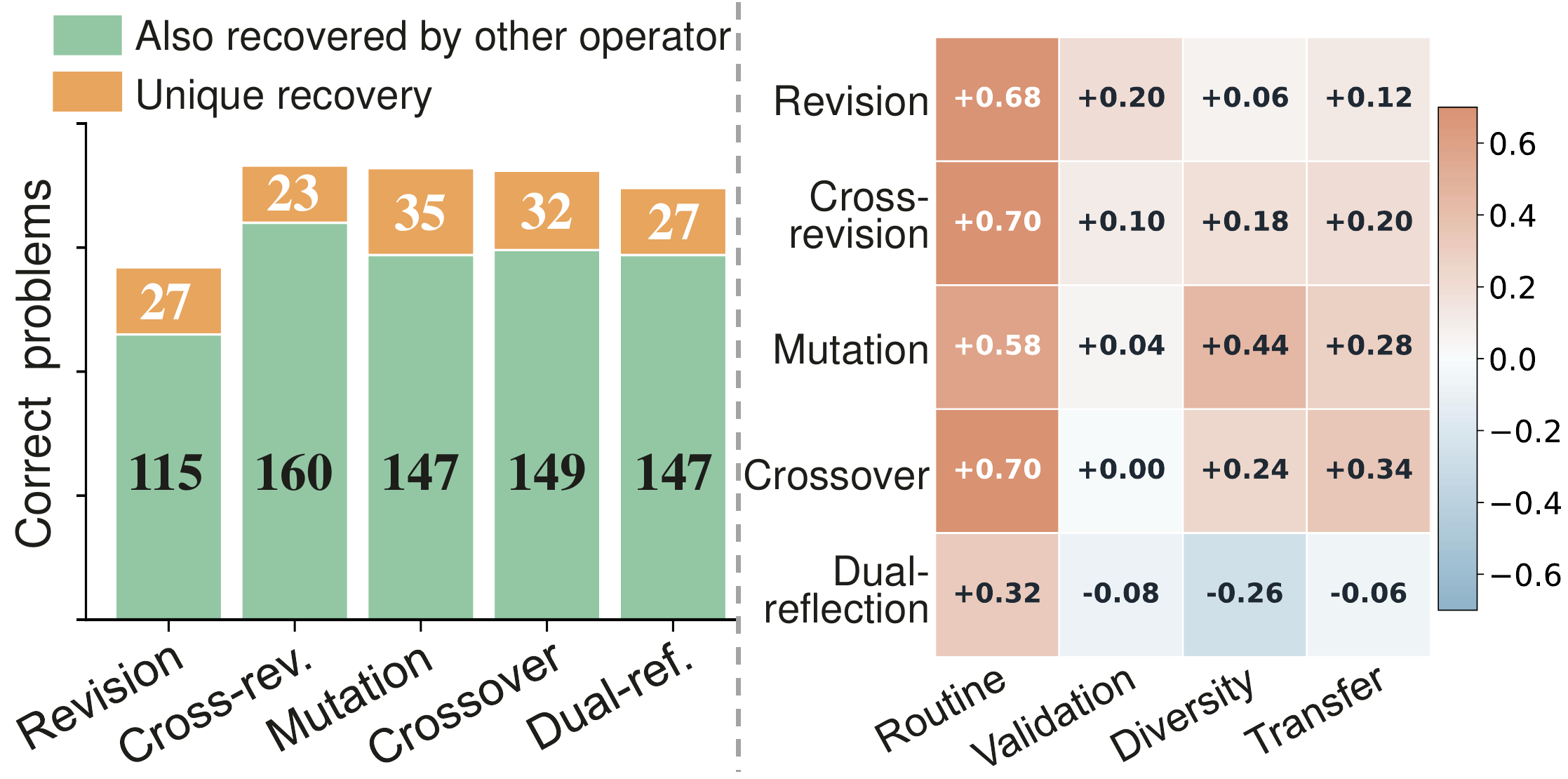}
  \caption{Diagnostics motivating multi-operator skill evolution.
  The left panel separates shared and operator-unique recoveries among
  failures of the initialized bank.
  The right panel reports judge-scored changes in retrieved skill
  guidance after evolution.}
  \label{fig:evolution-motivation}
\end{figure}

The initialized bank leaves hard problems unsolved even when the solver
retrieves relevant skills. \method uses these failures to evolve
existing skills through distinct genetic-inspired repair operators.
Figure~\ref{fig:evolution-motivation} shows that the operators uniquely
recover different failures and that evolution changes the quality of
retrieved guidance. These findings motivate both multi-operator
evolution and bank fusion; the diagnostic protocol appears in
Appendix~\ref{sec:supp-motivation-diagnostics}.

\section{Methodology}
\label{sec:methodology}


\method trains a dedicated skill evolver to transform retrieved parent
skills on hard negatives. Downstream rewards from the fixed solver both
train the evolver and select child skills for bank write-back. Separate
operator-specific runs produce banks that are fused for inference.




\subsection{Hard Negative Mining}
\label{sec:hard-negative-mining}

We construct the hard-negative pool once from the initialization
rollouts described in the \emph{Problem Setup} subsection.
A hard negative is a problem whose initial base-solver rollout receives
an evaluator reward below a task-specific success threshold.
Formally, we define
\begin{equation}
\begin{aligned}
  \mathcal{H}
  = \bigl\{x_i \;:\;&
  (x_i,a_0(x_i),r_0(x_i))\in\mathcal{T}_0,\\
  & r_0(x_i)<\tau
  \bigr\},
\end{aligned}
\end{equation}
where $\tau$ denotes the success threshold.
For math, $r_0(x_i)$ is determined by the benchmark answer checker;
for code, it is derived from executable tests.


The hard-negative pools are fixed before skill evolution begins.
At evolution step $t$, each operator-specific run samples a target
batch $\mathcal{B}_t\subseteq\mathcal{H}$ and retrieves
parent and support skills from its current skill bank.
These sampled hard negatives serve as evolution targets for the
framework.

\subsection{Skill Evolution Framework}
\label{sec:skill-evolution-framework}

The core of \method is a coupled process that evolves the skill bank
while simultaneously training a dedicated skill evolver through
reinforcement learning.
Evolution is conducted through five separate operator-specific runs.
The formulation below describes one such run; because the operator is
fixed throughout the run, we suppress operator and domain indices.
Each run maintains an evolving skill bank $\mathcal{S}_t$ and a
trainable skill evolver $q_{\phi_t}$.
The evolver parameters $\phi_t$ are updated throughout training.
The base solver parameters $\theta$ remain fixed.

For each target problem
$x\in\mathcal{B}_t\subseteq\mathcal{H}$, the run retrieves
parent and support skills
$\mathcal{R}_t(x)$ from its current bank.
The evolver then applies the run's operator to generate $K$ candidate
child skills:
\begin{equation}
  s_{x,j}^{\prime}
  \sim
  q_{\phi_t}
  \bigl(
    \cdot
    \mid x,\mathcal{R}_t(x)
  \bigr),
  \qquad j=1,\ldots,K .
\end{equation}
The candidate set for target $x$ is
\begin{equation}
  \mathcal{C}_t(x)
  =
  \{s_{x,j}^{\prime}\}_{j=1}^{K}.
\end{equation}

Each candidate child skill is evaluated by conditioning the fixed base
solver on that skill and measuring its downstream task performance.
The evaluation can include a
small set $\mathcal{N}(x)$ of related neighboring problems, with their
scores aggregated into the candidate reward
$R_{x,j}$.
This reward therefore reflects whether a child skill provides useful
guidance beyond only its source problem.

The candidate-level rewards serve two connected purposes.
First, they provide the reinforcement-learning signal used to train the
skill evolver to generate stronger child skills in subsequent
iterations.
Second, they guide the selection of beneficial child skills for
write-back into the skill bank.

We train the evolver using \grpo-style optimization
\citep{schulman2017ppo,ouyang2022training,shao2024deepseekmath,
yu2025dapo}.
For each target problem, its $K$ candidate child skills form a
comparison group, and we compute the group-relative advantage
\begin{equation}
  A_{x,j}
  =
  R_{x,j}
  -
  \frac{1}{K}
  \sum_{\ell=1}^{K}
  R_{x,\ell}.
\end{equation}
These relative advantages train the evolver to increase the likelihood
of transformations that produce more effective child skills.
Unlike conventional answer-level reinforcement learning, the optimized output here is a reusable skill that can subsequently support the fixed base solver.
In parallel, the downstream rewards determine which candidate child
skills are written back to the bank.
Let $\mathcal{W}_t$ denote the accepted children from the
current batch.
Within the write-back budget, we retain high-reward,
non-redundant candidates and update the operator-specific bank as
\begin{equation}
  \mathcal{S}_{t+1}
  =
  \mathrm{Prune}
  \bigl(
    \mathcal{S}_t
    \cup
    \mathcal{W}_t
  \bigr),
\end{equation}
where
$\mathrm{Prune}$ removes duplicate or low-priority entries.

Consequently, each iteration improves two components:
the skill bank acquires stronger child skills, while the evolver learns
to produce more effective skill transformations.



\subsection{Genetic-Inspired Evolution Operators}
\label{sec:evolution-operators}

Inspired by genetic evolution, \method treats existing skills as
parent material and transforms them into candidate child skills through
controlled inheritance, variation, and recombination.
Rather than relying on unconstrained reflection, each operator defines
a distinct transformation over the parent skills and their supporting
evidence.


Specifically, \textbf{Revision} conservatively repairs a weak strategy
while preserving its main solution route.
\textbf{Cross-revision} transfers a compatible motif or safeguard from
a related problem without replacing the target skill's backbone.
\textbf{Mutation} introduces a coherent alternative route, typically
changing the underlying strategy or algorithmic family.
\textbf{Crossover} preserves a strong backbone from one parent and
incorporates a compatible local improvement from another.
Finally, \textbf{Dual-reflection} strengthens correctness-critical
safeguards while keeping the main route largely stable.
Together, these operators generate complementary descendants that
address different forms of skill failure.

All operators follow a shared interface: they select
operator-specific support evidence, transform one or more parent skills
into a candidate child skill, and produce the guidance used for
downstream evaluation.
Detailed support-selection rules and prompt templates are provided in
Appendix~\ref{sec:supp-operator-details}.

\begin{figure}[t]
  \centering
  \includegraphics[width=\linewidth]{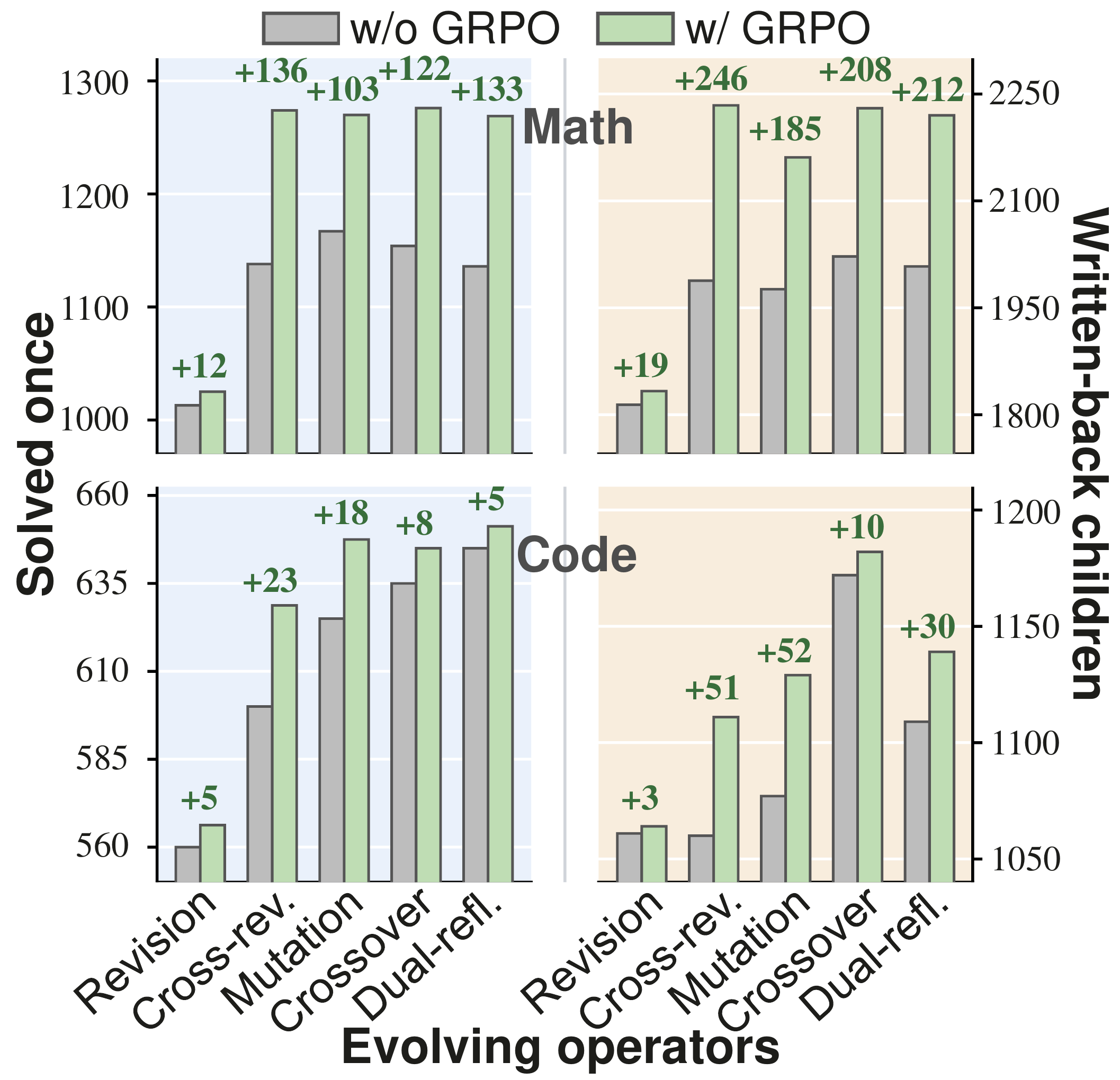}
  \caption{Effect of RL training on skill-evolution efficiency.
    \emph{Solved once} counts hard negatives solved at least once during
    evolution, while \emph{written-back children} counts child skills
    accepted into the bank.}
  \label{fig:rl-efficiency}
\end{figure}

\begin{figure}[t]
  \centering
  \includegraphics[width=\linewidth]{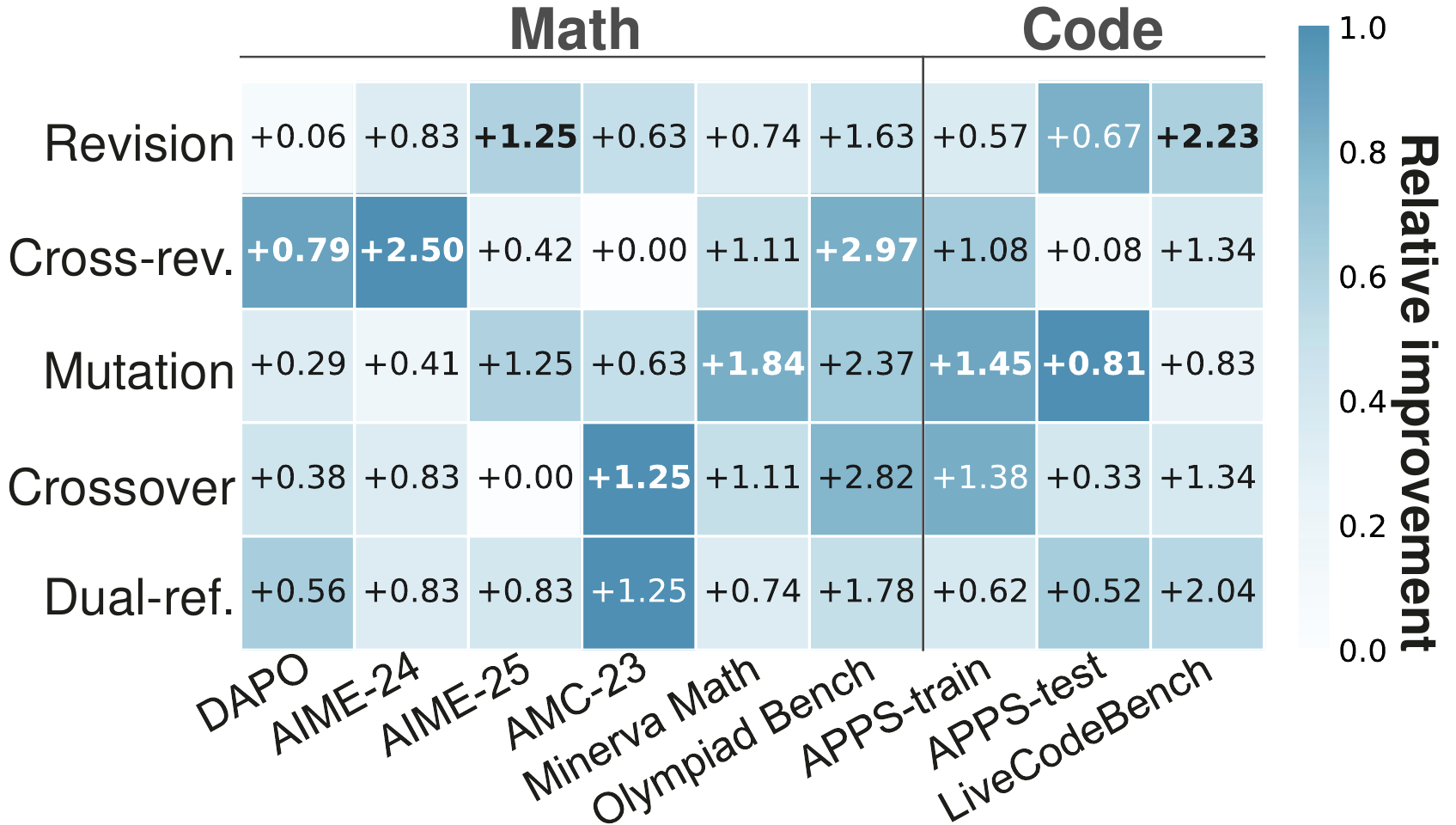}
  \caption{Operator-level comparison for the five skill evolution operators. The operators emphasize different support metrics, explaining why Complementary Skill Bank Fusion can outperform reliance on a single operator.}
  \label{fig:operator-advantages}
\end{figure}

\subsection{Complementary Skill Bank Fusion}
\label{sec:skill-bank-fusion}

The operator-specific skill banks capture complementary repair
patterns: different operators recover different subsets of failures
and produce distinct forms of skill guidance.
We therefore introduce \emph{Complementary Skill Bank Fusion}, which
consolidates the five operator-specific banks into a single evolved
skill bank under the same overall bank-size budget.

Let $\mathcal{O}$ denote the set of five evolution operators.
Given operator-specific banks
$\{\mathcal{S}^{(o)}\}_{o\in\mathcal{O}}$ and a target bank size $B$,
the fusion procedure selects
\begin{equation}
  \mathcal{S}^{\mathrm{mix}}
  \subseteq
  \bigcup_{o\in\mathcal{O}}\mathcal{S}^{(o)},
  \qquad
  |\mathcal{S}^{\mathrm{mix}}|\le B .
\end{equation}
Rather than simply pooling all evolved skills, the procedure balances
problem coverage, source reliability, operator complementarity, and
embedding diversity, while removing duplicate or low-priority records.
This allows the fused bank to inherit complementary strengths without
increasing the overall skill-bank budget.

We refer to this configuration as \emph{Fusion}; details are provided in Appendix~\ref{sec:supp-fusion}.

\subsection{Procedural Skill Interface}
\label{sec:procedural-skill-interface}

Skills evolved against specific hard negatives may retain
source-specific details that are useful for the source problem but less
suitable for retrieval on new problems.
After skill-bank fusion, the Procedural Skill Interface converts the
selected evolved skills into compact, retrieval-facing procedural
skills and matches them to target problems at inference time.
As summarized in Algorithm~\ref{alg:psi}, the interface consists of
three stages: skill abstraction, query matching, and solver-facing
composition.

This interface separates source-specific skill evolution from
retrieval-time reuse: evolution can exploit detailed feedback from hard
negatives, whereas the solver receives compact and reusable guidance at
inference time.
The prompt templates and filtering rules are provided in
Appendix~\ref{sec:supp-psi}.



\begin{table*}[t]
  \centering
  \setlength{\tabcolsep}{0.5pt}
  \begin{tabular}{@{}>{\centering\arraybackslash}m{0.065\textwidth}
                  >{\centering\arraybackslash}m{0.110\textwidth}
                  *{7}{>{\centering\arraybackslash}m{0.110\textwidth}}@{}}
    \toprule
    & & \multicolumn{5}{c}{\textbf{Math}}
    & \multicolumn{2}{c}{\textbf{Code}} \\
    \cmidrule(lr){3-7} \cmidrule(lr){8-9}
    \tablehead{\textbf{Model}} & \tablehead{\textbf{Method}}
    & \tablehead{\textbf{AIME}\\\textbf{24}}
    & \tablehead{\textbf{AIME}\\\textbf{25}}
    & \tablehead{\textbf{AMC}\\\textbf{23}}
    & \tablehead{\makebox[0pt][c]{\textbf{Minerva}}\\\textbf{Math}}
    & \tablehead{\makebox[0pt][c]{\textbf{Olymp.}}\\\makebox[0pt][c]{\textbf{Bench}}}
    & \tablehead{\textbf{APPS-}\\\textbf{test}}
    & \tablehead{\makebox[0pt][c]{\textbf{LiveCode}}\\\makebox[0pt][c]{\textbf{Bench}}} \\
    \midrule
    \multirow{12}{*}{\rotatebox[origin=c]{90}{Qwen3-4B-Instruct}} & No skill & 53.75 & 42.08 & 87.50 & 16.18 & 40.65 & 36.90 & 41.98 \\
    & Traj RAG & 57.08 & 40.83 & 91.56 & 29.04 & 54.15 & 28.58 & 31.17 \\
    & Reflexion & 57.92 & 42.92 & 93.12 & 32.35 & 66.77 & 41.82 & 42.18 \\
    & ExpeL-SI & 59.17 & 41.67 & 92.50 & 30.88 & 66.32 & 41.64 & 39.63 \\
    \cmidrule(lr){2-9}
    & \multicolumn{8}{c}{\emph{Tiny Skill Bank}} \\
    \cmidrule(lr){2-9}
    & Init & 60.42 & 45.00 & 91.56 & 33.82 & 66.32 & 44.31 & 41.92 \\
    & \ourscell{Single-op} & \ourscell{61.50} & \ourscell{45.75} & \ourscell{92.31} & \ourscell{34.93} & \ourscell{68.63} & \ourscell{44.79} & \ourscell{43.44} \\
    & \ourscell{Fusion} & \ourscell{\textbf{62.92}} & \ourscell{\textbf{47.08}} & \ourscell{\text{92.81}} & \ourscell{\textbf{36.03}} & \ourscell{\textbf{69.88}} & \ourscell{\textbf{45.08}} & \ourscell{\textbf{45.55}} \\
    \cmidrule(lr){2-9}
    & \multicolumn{8}{c}{\emph{Large Skill Bank}} \\
    \cmidrule(lr){2-9}
    & Init & 61.67 & 46.67 & 92.19 & 37.13 & 67.36 & 44.38 & 44.02 \\
    & \ourscell{Single-op} & \ourscell{66.67} & \ourscell{50.42} & \ourscell{93.44} & \ourscell{37.13} & \ourscell{67.95} & \ourscell{44.75} & \ourscell{44.78} \\
    & \ourscell{Fusion} & \ourscell{\textbf{67.08}} & \ourscell{\textbf{51.67}} & \ourscell{\textbf{94.38}} & \ourscell{\textbf{38.24}} & \ourscell{\textbf{68.40}} & \ourscell{\textbf{44.92}} & \ourscell{\textbf{45.17}} \\
    \midrule
    \multirow{10}{*}{\rotatebox[origin=c]{90}{GPT Family}} & No skill & 44.17 & 42.50 & 79.38 & 25.37 & 45.99 & 52.51 & 41.54 \\
    & Traj RAG & 50.42 & 44.68 & 85.32 & 32.35 & 61.35 & 41.85 & 50.76 \\
    & Reflexion & 50.42 & 46.54 & 86.12 & 31.99 & 60.83 & 42.23 & 46.44 \\
    & ExpeL-SI & 47.92 & 45.83 & 86.57 & 33.89 & 59.94 & 52.63 & 47.01 \\
    \cmidrule(lr){2-9}
    & \multicolumn{8}{c}{\emph{Tiny Skill Bank}} \\
    \cmidrule(lr){2-9}
    & Init & 48.75 & 42.92 & 86.25 & 33.46 & 61.72 & 53.80 & 52.80 \\
    & \ourscell{Fusion} & \ourscell{\textbf{50.83}} & \ourscell{\text{43.75}} & \ourscell{\textbf{86.88}} & \ourscell{\textbf{34.12}} & \ourscell{\textbf{62.02}} & \ourscell{\textbf{53.85}} & \ourscell{\textbf{53.50}} \\
    \cmidrule(lr){2-9}
    & \multicolumn{8}{c}{\emph{Large Skill Bank}} \\
    \cmidrule(lr){2-9}
    & Init & 48.75 & 43.75 & 88.75 & 34.56 & 60.24 & 53.98 & 53.94 \\
    & \ourscell{Fusion} & \ourscell{\textbf{52.08}} & \ourscell{\textbf{49.17}} & \ourscell{\textbf{89.69}} & \ourscell{\textbf{35.29}} & \ourscell{\textbf{61.57}} & \ourscell{\textbf{54.13}} & \ourscell{\textbf{55.53}} \\
    \bottomrule
  \end{tabular}
\caption{Main results on math and code benchmarks.
Math columns report accuracy (\%), while code columns report mean
unit-test pass rate (\%).
Single-op denotes the arithmetic mean across the five independently
evolved operator-specific banks, while Fusion denotes Complementary
Skill Bank Fusion under the same overall bank-size budget.
Shaded rows indicate our evolved skill-bank variants, and bold values
mark the best result within each backbone and skill-bank-size block.
For the GPT Family rows, math uses GPT-4.1 mini and code uses GPT-4o mini.}
  \label{tab:main-results}
\end{table*}



\section{Experiments}
\label{sec:experiments}

We organize our evaluation around four research questions.
\begin{itemize}
  \item \textbf{RQ1: Performance.}
  Does direct skill evolution improve over initialized skill banks and
  non-direct-evolution baselines across math and code?

  \item \textbf{RQ2: Fusion.}
  Under the same overall bank-size budget, does fusion outperform
  average single-operator evolution on held-out benchmarks?

  \item \textbf{RQ3: RL efficiency.}
  Does RL-based training of the skill evolver improve evolution
  efficiency under a comparable write-back budget?

  \item \textbf{RQ4: Proceduralization.}
  Does the Procedural Skill Interface improve the retrieval and reuse
  quality of evolved skills on unseen problems?
\end{itemize}

\subsection{Experimental Setup}

\paragraph{Evolution targets and evaluation benchmarks.}
Separate problem sets are used for skill evolution and held-out
evaluation.
Fixed hard-negative pools are constructed from problems on which the
base solver fails during skill-bank initialization.
This procedure yields approximately 4,000 math problems from DAPO
\citep{yu2025dapo} and 2,200 code problems from the APPS training set
\citep{hendrycks2021apps}.
These hard-negative pools serve as evolution targets and support
in-domain diagnostic analyses.

The resulting skill banks are evaluated on separate held-out
benchmarks.
The math evaluation suite comprises AIME24, AIME25, AMC23, Minerva
Math, and OlympiadBench
\citep{maa2024aime,maa2025aime,maa2023amc,lewkowycz2022minerva,
he2024olympiadbench}.
The code evaluation suite comprises APPS-test and LiveCodeBench
\citep{hendrycks2021apps,jain2024livecodebench}.
Accuracy is reported for math, while mean unit-test pass rate is
reported for code.

\paragraph{Baselines.}
The comparison includes representative paradigms spanning no external
support, trajectory reuse, experience-derived reflection, and static
skill-bank retrieval.
\emph{No skill} evaluates the solver without retrieved skill support,
whereas \emph{Traj RAG} retrieves relevant trajectory-derived guidance.
\emph{Reflexion} generates verbal feedback from prior failures, and
\emph{ExpeL-SI} extracts reusable skills from accumulated experience
\citep{NEURIPS2023_1b44b878,zhao2024expel}.
\emph{Init} uses the initialized skill bank without subsequent
evolution.
Among the proposed evolution settings, \emph{Single-op} reports the
arithmetic mean across the five independently evolved
operator-specific banks, while \emph{Fusion} consolidates these banks under the same overall
bank-size budget.

The representative baselines are adapted to the common fixed-solver,
retrieval, and evaluation protocol; implementation details appear in
Appendix~\ref{sec:supp-experimental-details}.


\paragraph{Backbones and implementation.}
Qwen3-4B-Instruct serves as both the primary open-weight solver and the
initialization for the separately trained skill evolver; the solver
remains frozen during RL training.
GPT-family experiments use GPT-4.1 mini for math and GPT-4o mini for
code.
Tiny banks contain 500 initialized skills, whereas large banks use the
full curated bank for each domain.
Within each setting, all methods share the same solver, retriever,
evaluator, prompts, decoding configuration, and metrics.
Repeated-run statistics, seed settings, and implementation details are provided in Appendix~\ref{sec:supp-experimental-details}.

\subsection{Main Results}
\label{sec:main-results}

Table~\ref{tab:main-results} supports two main takeaways.


\noindent\textbf{Takeaway \ding{182}: Direct skill evolution consistently improves initialized skill banks.}
Across both bank sizes and both solver families, Fusion outperforms the
corresponding Init setting on every reported math and code benchmark.
For Qwen3-4B-Instruct, the improvements are particularly pronounced on
AIME24 and AIME25.
The same consistent trend holds for the GPT-family solvers, showing
that the evolved skill banks remain effective beyond the primary
open-weight solver setting.


\noindent\textbf{Takeaway \ding{183}: Complementary Skill Bank Fusion improves over average single-operator evolution.}
For Qwen3-4B-Instruct, Fusion outperforms Single-op on every benchmark
under both the tiny- and large-bank settings.
This consistent advantage supports consolidating complementary
operator-specific banks under a fixed overall budget.





\subsection{Efficiency Analysis}

\noindent\textbf{Takeaway \ding{184}: RL training improves skill-evolution efficiency without simply increasing write-back volume.}
Figure~\ref{fig:rl-efficiency} compares each operator with and without
RL training of the skill evolver.
\emph{Solved once} denotes the number of hard negatives solved at least
once during evolution, while \emph{written-back children} denotes the
number of child skills accepted into the bank.
Across both math and code, RL-trained evolvers recover more hard
negatives under comparable write-back budgets, indicating that RL
enables the evolver to generate more effective child skills.



\subsection{Operator Analysis}

\noindent\textbf{Takeaway \ding{185}: The five genetic-inspired
operators exhibit complementary strengths.}
Figure~\ref{fig:evolution-motivation} shows operator-unique recoveries.
Figure~\ref{fig:operator-advantages} and
Appendix~\ref{sec:supp-operator-results} provide the full comparisons.
Fusion combines these complementary strengths under the same bank-size
budget (Table~\ref{tab:main-results}).



\subsection{Proceduralization Analysis}

\begin{figure}[t]
  \centering
  \includegraphics[width=\linewidth]{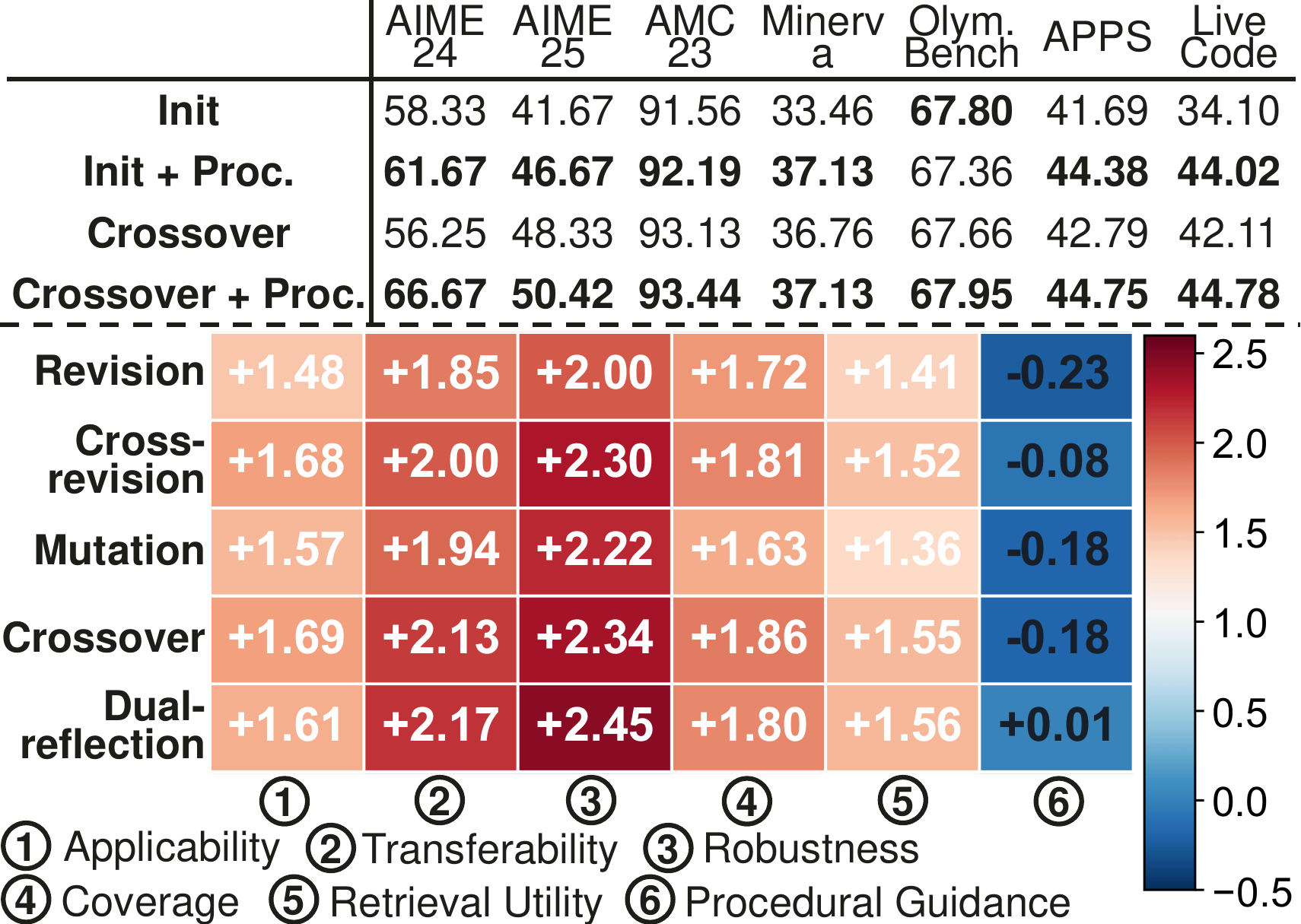}
  \caption{Effect of proceduralization on downstream performance and
  skill-interface quality. Top: performance before and after
  proceduralization for the initialized and Crossover skill banks. Bottom: paired-judge score
  differences between proceduralized skills and source skills across six interface-quality dimensions; positive
  values indicate improvement.}
  \label{fig:proceduralization-analysis}
\end{figure}

\noindent\textbf{Takeaway \ding{186}: The Procedural Skill Interface
improves both downstream performance and retrieval-facing skill
quality.}
Figure~\ref{fig:proceduralization-analysis} shows that
proceduralization improves the initialized bank on six of seven
benchmarks and the Crossover bank on all seven. Paired judgments of
all five operators' outputs show gains in five interface-quality
dimensions, with smaller changes in procedural guidance. The protocol appears in
Appendix~\ref{sec:supp-proceduralization-judge}.

\section{Conclusion}

We presented \method, an RL-driven framework that directly evolves
reusable skills while keeping the base solver fixed.
Given hard-negative problems, a dedicated skill evolver transforms
existing parent skills into candidate child skills through
genetic-inspired operators, with downstream rewards jointly training
the evolver and guiding selective write-back.
Complementary Skill Bank Fusion consolidates operator-specific banks,
while the Procedural Skill Interface supports retrieval-time reuse.
Experiments across math and code show consistent improvements over
initialized skill banks under both Qwen3-4B-Instruct and GPT-family
solvers, with analyses confirming the complementary strengths of the
operators and the efficiency gains from RL training.

\bibliography{custom}

\clearpage
\appendix
\section{Further Method and Operator Analysis}

\label{sec:additional-analysis}

\begin{mainalgorithm}{Procedural Skill Interface}
  \label{alg:psi}

  1: & \textbf{Input:}
  evolved skill bank $\mathcal{S}^{\mathrm{evo}}$,
  target problem $x$, retriever $R$.
  \textbf{Output:} solver support $Z(x)$. \\

  2: & \textbf{Skill abstraction:}
  convert each evolved skill into one or more reusable procedural
  skills. \\

  3: & $\displaystyle
  \mathcal{S}^{\mathrm{proc}}
  \leftarrow
  \bigcup_{s\in\mathcal{S}^{\mathrm{evo}}}
  \mathrm{Proceduralize}(s).$ \\

  4: & \textbf{Query matching:} \\

  5: & Extract problem keys:
  $Q(x)\leftarrow\mathrm{KeyExtract}(x)$. \\

  6: & Retrieve procedural skills: \newline
  $U_x\leftarrow
  R\!\left(Q(x),\mathcal{S}^{\mathrm{proc}}\right)$. \\

  7: & Filter retrieved skills by relevance, diversity, and
  reliability:
  $\widehat{U}_x\leftarrow\mathrm{Filter}(U_x)$. \\

  8: & \textbf{Support composition:}
  $Z(x)\leftarrow\mathrm{Compose}(\widehat{U}_x)$. \\

  9: & \textbf{return} $Z(x)$. \\

\end{mainalgorithm}

\subsection{Operator-Level Interpretation}

Figure~\ref{fig:operator-advantages} shows that the relative gains of
different operators vary across benchmarks, consistent with the
operator-unique recoveries observed in
Figure~\ref{fig:evolution-motivation}.
Detailed results in the Supplementary Material further show that no
single operator consistently dominates across all benchmarks.
This complementarity motivates Complementary Skill Bank Fusion, which
consolidates the distinct strengths of the operator-specific banks.
In Table~\ref{tab:main-results}, \emph{Single-op} reports the
arithmetic mean across the five independently evolved banks, whereas
\emph{Fusion} measures the benefit of consolidating them under the same
overall bank-size budget.

\subsection{Baseline Comparability}

Several recent skill-learning systems are tightly coupled to specific
agent environments, skill representations, or jointly optimized
solving policies, making their complete pipelines unsuitable for direct
comparison in the fixed-solver math and code setting. Accordingly,
benchmark-adapted instantiations of representative mechanisms are
implemented under a common protocol. Traj RAG captures
trajectory-level reuse, Reflexion captures failure-driven reflection,
and ExpeL-SI captures experience-to-skill induction. These
implementations isolate the core mechanism of each paradigm while
controlling for the solver, retrieval, and evaluation settings.


\section{Additional Experimental Details}
\label{sec:supp-experimental-details}

\subsection{Motivation Diagnostic Setup}
\label{sec:supp-motivation-diagnostics}

The motivation diagnostic reported in the main paper uses the math-side
DAPO-hard pool mined from failed initialization rollouts. We first retain the
six-way intersection of problem identifiers with results under the initialized
bank and all five operator-specific evolved banks: Revision, Cross-revision,
Mutation, Crossover, and Dual-reflection. The diagnostic set consists of the
first 1,000 problem identifiers in this retained pool. All six settings use the
same solver, retrieval protocol, and evaluator, so differences in the
diagnostic are attributable to the retrieved skills rather than changes in the
test set or solver configuration.

\paragraph{Operator complementarity.}
The operator-complementarity panel in the main paper measures which initialized failures are recovered by each evolved skill bank. Let
\begin{equation}
  \mathcal{F}_{\mathrm{init}}
  = \{q : \text{the initialized skill bank fails on } q\}
\end{equation}
be the subset of the shared evaluation set missed by the initialized bank. For
an operator $o$, let $\mathrm{Solve}_o(q)$ indicate whether the bank evolved by
$o$ solves $q$. We define the recovered set as
\begin{equation}
  \mathcal{R}_o
  = \{q\in\mathcal{F}_{\mathrm{init}} : \mathrm{Solve}_o(q)=1\}.
\end{equation}
We decompose $\mathcal{R}_o$ into shared and unique recoveries:
\begin{equation}
\begin{aligned}
  \mathcal{R}^{\mathrm{shared}}_o
  &= \{q\in\mathcal{R}_o : \exists o'\neq o,\; q\in\mathcal{R}_{o'}\},\\
  \mathcal{R}^{\mathrm{unique}}_o
  &= \mathcal{R}_o \setminus \mathcal{R}^{\mathrm{shared}}_o .
\end{aligned}
\end{equation}
The bar height is $|\mathcal{R}_o|$, while the unique segment is
$|\mathcal{R}^{\mathrm{unique}}_o|$. This analysis tests whether the operators
simply recover the same failures or provide complementary repair behavior.

\paragraph{Retrieved-skill repair signature.}
The retrieved-skill repair-signature panel in the main paper studies how
retrieved skills change on flip cases where the initialized solver fails and
the operator-specific solver succeeds. For each operator, we independently
shuffle its flip set with random seed 29 and retain at most 50 cases. For each
case, we collect the retrieved skill context shown to the solver from the
initialized bank and from the operator-specific evolved bank. The
\texttt{openai/gpt-5.4} judge is queried with temperature $0.0$ and scores each
retrieved context on a 1--5 scale along routine support, validation support,
route diversity, and transfer support. Each score is the four-call mean defined
in Section~\ref{sec:supp-repeated-evaluation}. Routine support measures
whether the retrieved skills provide executable steps; validation support
measures whether they provide checking mechanisms; route diversity measures
whether they give complementary solution routes rather than repeated hints;
and transfer support measures whether the guidance abstracts beyond the source
problem surface.

For operator $o$ and dimension $d$, the plotted value is
\begin{equation}
  \Delta_{o,d}
  = \frac{1}{|\mathcal{Q}_o|}\sum_{q\in\mathcal{Q}_o}
  \left[J_d(Z_o(q)) - J_d(Z_{\mathrm{init}}(q))\right],
\end{equation}
where $\mathcal{Q}_o$ is the sampled case set, $Z_{\mathrm{init}}(q)$ and
$Z_o(q)$ are the retrieved contexts from the initialized and evolved banks,
and $J_d(\cdot)$ is the judge score for dimension $d$.

\subsection{Proceduralization Judge Study}
\label{sec:supp-proceduralization-judge}

The proceduralization analysis in Figure~\ref{fig:proceduralization-analysis} of the main paper uses paired
comparisons between source skills and the proceduralized units produced from
them. We evaluate outputs from the five operator-specific math skill banks:
Revision, Cross-revision, Mutation, Crossover, and Dual-reflection. Each
proceduralized unit retains identifiers for its source skill, problem, and
operator, enabling alignment with the source skill that produced it. When one
source skill yields multiple procedural units, those units are treated as one
proceduralized support group.

For each operator bank $b$, we shuffle the aligned pairs with sampling seed
$20260526$ and retain the first 100 pairs
\begin{equation}
  (s_{b,i}, U_{b,i}), \qquad i=1,\ldots,100,
\end{equation}
where $s_{b,i}$ is the source skill and $U_{b,i}$ is its proceduralized support
group. This gives $5\times100=500$ paired examples. The judge scores both sides
of each pair on a 1--5 scale along six dimensions:
\emph{applicability}, whether the skill states when it should be used;
\emph{transferability}, whether it abstracts a pattern applicable to unseen
related problems; \emph{robustness}, whether it remains reliable under
ambiguous or failure-prone conditions; \emph{coverage}, whether it captures
the essential conditions and solution components; \emph{retrieval utility},
whether it would help as top-$k$ retrieved support; and \emph{procedural
guidance}, whether it provides concrete, actionable solving guidance.

\paragraph{Judge configuration.}
We use \texttt{openai/gpt-5.4} with temperature $0.0$ and a maximum output
length of 900 tokens. Responses are constrained to JSON objects. Each request
has a 180-second timeout and permits up to three retries, and evaluation jobs
are distributed across eight shards. Each reported judge score uses the
four-call averaging protocol in Section~\ref{sec:supp-repeated-evaluation}.

For dimension $d$, we compute the paired improvement
\begin{equation}
  \Delta_{b,d}
  = \frac{1}{100}\sum_{i=1}^{100}
  \left[J_d(U_{b,i}) - J_d(s_{b,i})\right].
\end{equation}
The paired comparison shows consistent gains in applicability,
transferability, robustness, coverage, and retrieval utility. Procedural
guidance remains close to the source-skill level, indicating that the interface
primarily improves when and how skills can be retrieved and reused rather than
merely lengthening their procedural content.

\subsection{Benchmarks and Metrics}
\label{sec:supp-benchmarks}

Table~\ref{tab:supp-benchmark-details} summarizes the benchmark splits,
sources, and evaluation sizes. Math benchmarks use answer-level accuracy after
benchmark-specific normalization, while code benchmarks use mean unit-test pass
rate.

\paragraph{Computational budget.}
The primary open-weight backbone is Qwen3-4B-Instruct, a 4B-parameter model.
Each RL-driven skill-evolution training run uses four NVIDIA A100 GPUs for 24
hours. GPT-family mini models are
accessed through hosted APIs.

\paragraph{Experimental setup and hyperparameters.}
We use fixed experimental settings across methods within each backbone and
domain rather than tuning hyperparameters separately for each baseline. All
methods are evaluated with the same solver, retrieval protocol, evaluator,
benchmark prompts, and metric scripts. We do not perform hyperparameter search
on test benchmarks; skill-bank size, operator setting, write-back budget, and
evaluation protocol are fixed before the final evaluations.

\paragraph{Reporting convention and decoding parameters.}
Unless otherwise specified, decoding uses temperature $0.2$, top-$p$ $0.9$,
top-$k$ $20$, and a maximum generation length of $16{,}384$ tokens.

\subsection{Repeated Evaluation and Judge Aggregation}
\label{sec:supp-repeated-evaluation}

\paragraph{Benchmark evaluation.}
For every fixed solver--skill-bank configuration and benchmark, we conduct
eight independent evaluation runs. Each run evaluates the full benchmark once
and samples one solver completion per problem. We use inference seeds
$\{2000,2001,\ldots,2007\}$ for all methods compared within a
backbone--domain setting, enabling paired comparisons under matched sampling
conditions. If
$M^{(r)}$ is the benchmark-level metric from run $r$, the reported value is
\begin{equation}
  \overline{M}=\frac{1}{8}\sum_{r=1}^{8}M^{(r)}.
\end{equation}
These repetitions measure evaluation-time stochasticity; the skill evolver and
skill bank are not retrained between evaluation runs.

Tables~\ref{tab:supp-qwen-main-std}--\ref{tab:supp-gpt-main-code-std}
report $\overline{M}\pm\sigma_M$, where $\sigma_M$ is the empirical standard
deviation of the eight run-level metrics. All values are percentage points.

\begin{table*}[tbp]
  \centering
  \setlength{\tabcolsep}{3pt}
  \begin{tabular}{@{}llccccc@{}}
    \toprule
    \textbf{Skill bank} & \textbf{Method} & \textbf{AIME 24} & \textbf{AIME 25} & \textbf{AMC 23} & \textbf{Minerva Math} & \textbf{OlympiadBench} \\
    \midrule
    None & No skill & $53.75{\pm}6.53$ & $42.08{\pm}6.41$ & $87.50{\pm}3.27$ & $16.18{\pm}1.35$ & $40.65{\pm}1.07$ \\
     & Traj RAG & $57.08{\pm}6.53$ & $40.83{\pm}6.36$ & $91.56{\pm}2.97$ & $29.04{\pm}1.66$ & $54.15{\pm}1.08$ \\
     & Reflexion & $57.92{\pm}6.41$ & $42.92{\pm}6.53$ & $93.12{\pm}2.59$ & $32.35{\pm}1.71$ & $66.77{\pm}1.03$ \\
     & ExpeL-SI & $59.17{\pm}6.36$ & $41.67{\pm}6.42$ & $92.50{\pm}2.67$ & $30.88{\pm}1.67$ & $66.32{\pm}1.02$ \\
    \midrule
    Tiny & Init & $60.42{\pm}6.28$ & $45.00{\pm}6.42$ & $91.56{\pm}2.97$ & $33.82{\pm}1.72$ & $66.32{\pm}1.03$ \\
     & Single-op & $61.50{\pm}6.42$ & $45.75{\pm}6.61$ & $92.31{\pm}2.81$ & $34.93{\pm}1.75$ & $68.63{\pm}1.00$ \\
     & Fusion & $62.92{\pm}6.28$ & $47.08{\pm}6.53$ & $92.81{\pm}2.81$ & $36.03{\pm}1.76$ & $69.88{\pm}1.00$ \\
    \midrule
    Large & Init & $61.67{\pm}6.42$ & $46.67{\pm}6.42$ & $92.19{\pm}2.81$ & $37.13{\pm}1.78$ & $67.36{\pm}1.02$ \\
     & Single-op & $66.67{\pm}6.17$ & $50.42{\pm}6.53$ & $93.44{\pm}2.65$ & $37.13{\pm}1.76$ & $67.95{\pm}1.01$ \\
     & Fusion & $67.08{\pm}6.28$ & $51.67{\pm}6.55$ & $94.38{\pm}2.59$ & $38.24{\pm}1.79$ & $68.40{\pm}1.02$ \\
    \bottomrule
  \end{tabular}
  \caption{Repeated-evaluation results for the Qwen3-4B-Instruct math benchmarks. Values are mean $\pm$ standard deviation over eight inference seeds; the corresponding means appear in Table~\ref{tab:main-results}.}
  \label{tab:supp-qwen-main-std}
\end{table*}

\begin{table*}[tbp]
  \centering
  \setlength{\tabcolsep}{3pt}
  \begin{tabular}{@{}llcc@{}}
    \toprule
    \textbf{Skill bank} & \textbf{Method} & \textbf{APPS-test} & \textbf{LiveCodeBench} \\
    \midrule
    None & No skill & $36.90{\pm}0.43$ & $41.98{\pm}3.17$ \\
     & Traj RAG & $28.58{\pm}0.41$ & $31.17{\pm}2.97$ \\
     & Reflexion & $41.82{\pm}0.45$ & $42.18{\pm}3.22$ \\
     & ExpeL-SI & $41.64{\pm}0.45$ & $39.63{\pm}3.14$ \\
    \midrule
    Tiny & Init & $44.31{\pm}0.45$ & $41.92{\pm}3.17$ \\
     & Single-op & $44.79{\pm}0.45$ & $43.44{\pm}3.20$ \\
     & Fusion & $45.08{\pm}0.45$ & $45.55{\pm}3.21$ \\
    \midrule
    Large & Init & $44.38{\pm}0.45$ & $44.02{\pm}3.21$ \\
     & Single-op & $44.75{\pm}0.45$ & $44.78{\pm}3.19$ \\
     & Fusion & $44.92{\pm}0.45$ & $45.17{\pm}3.20$ \\
    \bottomrule
  \end{tabular}
  \caption{Repeated-evaluation results for the Qwen3-4B-Instruct code benchmarks. Values are mean $\pm$ standard deviation over eight inference seeds; the corresponding means appear in Table~\ref{tab:main-results}.}
  \label{tab:supp-qwen-main-code-std}
\end{table*}

\begin{table*}[tbp]
  \centering
  \setlength{\tabcolsep}{3pt}
  \begin{tabular}{@{}llccccc@{}}
    \toprule
    \textbf{Skill bank} & \textbf{Method} & \textbf{AIME 24} & \textbf{AIME 25} & \textbf{AMC 23} & \textbf{Minerva Math} & \textbf{OlympiadBench} \\
    \midrule
    None & No skill & $44.17{\pm}6.61$ & $42.50{\pm}6.61$ & $79.38{\pm}4.17$ & $25.37{\pm}1.60$ & $45.99{\pm}1.08$ \\
     & Traj RAG & $50.42{\pm}6.53$ & $44.68{\pm}6.41$ & $85.32{\pm}3.64$ & $32.35{\pm}1.71$ & $61.35{\pm}1.06$ \\
     & Reflexion & $50.42{\pm}6.53$ & $46.54{\pm}6.67$ & $86.12{\pm}3.54$ & $31.99{\pm}1.71$ & $60.83{\pm}1.06$ \\
     & ExpeL-SI & $47.92{\pm}6.41$ & $45.83{\pm}6.61$ & $86.57{\pm}3.52$ & $33.89{\pm}1.73$ & $59.94{\pm}1.06$ \\
    \midrule
    Tiny & Init & $48.75{\pm}6.65$ & $42.92{\pm}6.53$ & $86.25{\pm}3.54$ & $33.46{\pm}1.74$ & $61.72{\pm}1.06$ \\
     & Fusion & $50.83{\pm}6.61$ & $43.75{\pm}6.53$ & $86.88{\pm}3.47$ & $34.12{\pm}1.75$ & $62.02{\pm}1.06$ \\
    \midrule
    Large & Init & $48.75{\pm}6.41$ & $43.75{\pm}6.53$ & $88.75{\pm}3.27$ & $34.56{\pm}1.72$ & $60.24{\pm}1.05$ \\
     & Fusion & $52.08{\pm}6.41$ & $49.17{\pm}6.61$ & $89.69{\pm}3.12$ & $35.29{\pm}1.75$ & $61.57{\pm}1.05$ \\
    \bottomrule
  \end{tabular}
  \caption{Repeated-evaluation results for the GPT-family mini math benchmarks. Values are mean $\pm$ standard deviation over eight inference seeds; the corresponding means appear in Table~\ref{tab:main-results}.}
  \label{tab:supp-gpt-main-std}
\end{table*}

\begin{table*}[tbp]
  \centering
  \setlength{\tabcolsep}{3pt}
  \begin{tabular}{@{}llcc@{}}
    \toprule
    \textbf{Skill bank} & \textbf{Method} & \textbf{APPS-test} & \textbf{LiveCodeBench} \\
    \midrule
    None & No skill & $52.51{\pm}0.45$ & $41.54{\pm}3.16$ \\
     & Traj RAG & $41.85{\pm}0.45$ & $50.76{\pm}3.21$ \\
     & Reflexion & $42.23{\pm}0.44$ & $46.44{\pm}3.21$ \\
     & ExpeL-SI & $52.63{\pm}0.45$ & $47.01{\pm}3.21$ \\
    \midrule
    Tiny & Init & $53.80{\pm}0.45$ & $52.80{\pm}3.20$ \\
     & Fusion & $53.85{\pm}0.45$ & $53.50{\pm}3.24$ \\
    \midrule
    Large & Init & $53.98{\pm}0.45$ & $53.94{\pm}3.23$ \\
     & Fusion & $54.13{\pm}0.45$ & $55.53{\pm}3.21$ \\
    \bottomrule
  \end{tabular}
  \caption{Repeated-evaluation results for the GPT-family mini code benchmarks. Values are mean $\pm$ standard deviation over eight inference seeds; the corresponding means appear in Table~\ref{tab:main-results}.}
  \label{tab:supp-gpt-main-code-std}
\end{table*}

\paragraph{Judge aggregation.}
For each item and scoring dimension, we perform four independent judge calls
with the same model, rubric, input, and decoding configuration. Let
$J_d^{(h)}(z)$ denote the score assigned to item $z$ on dimension $d$ by judge
call $h$. We first compute
\begin{equation}
  J_d(z)=\frac{1}{4}\sum_{h=1}^{4}J_d^{(h)}(z),
\end{equation}
and then aggregate these item-level means across examples and operators. Thus,
the judge quantities used in the diagnostic and proceduralization analyses are
four-call averages rather than single-call scores.

\paragraph{Artifact licenses and terms.}
We use publicly available benchmarks and model artifacts for research
evaluation and cite their original creators. We do not redistribute third-party
datasets or model checkpoints. Any released code or derived skill-bank
artifacts are intended for research use and will be distributed only in
accordance with the applicable licenses and terms of the underlying resources.

\begin{table*}[tbp]
  \centering
  \setlength{\tabcolsep}{3pt}
  \begin{tabular}{@{}>{\raggedright\arraybackslash}p{0.15\textwidth}
                  >{\raggedright\arraybackslash}p{0.11\textwidth}
                  >{\raggedright\arraybackslash}p{0.31\textwidth}
                  >{\centering\arraybackslash}p{0.08\textwidth}
                  >{\raggedright\arraybackslash}p{0.25\textwidth}@{}}
    \toprule
    \textbf{Benchmark} & \textbf{Domain} & \textbf{Split / source} &
    \textbf{Size} & \textbf{Use and metric} \\
    \midrule
    DAPO-hard & Math & Hard subset mined from the DAPO initialization pool after
    base-solver rollouts. & $\sim$4k & Evolution target pool; answer-level correctness. \\
    AIME24 & Math & Official 2024 AIME I and II problems \citep{maa2024aime}. &
    30 & Held-out math evaluation; answer-level correctness. \\
    AIME25 & Math & Official 2025 AIME I and II problems \citep{maa2025aime}. &
    30 & Held-out math evaluation; answer-level correctness. \\
    AMC23 & Math & 2023 AMC 12 A/B aggregate \citep{maa2023amc}. & 40 &
    Held-out math evaluation; answer-level correctness. \\
    Minerva Math & Math & Minerva quantitative-reasoning test set
    \citep{lewkowycz2022minerva}. & 272 & Held-out math evaluation;
    answer-level correctness. \\
    OlympiadBench & Math & Text-only open-ended math subset
    \citep{he2024olympiadbench}. & 675 & Out-of-distribution math evaluation;
    answer-level correctness. \\
    APPS-train-hard & Code & Hard subset mined from APPS training rollouts. &
    $\sim$2.2k & Evolution target pool; mean unit-test pass rate. \\
    APPS-test & Code & Official APPS test split \citep{hendrycks2021apps}. &
    5,000 & Held-out code evaluation; mean unit-test pass rate. \\
    LiveCodeBench & Code & 131-problem code-generation subset
    \citep{jain2024livecodebench}. & 131 & Out-of-distribution code evaluation;
    mean unit-test pass rate. \\
    \bottomrule
  \end{tabular}
  \caption{Benchmark splits, sources, and evaluation sizes. Internal hard
  subsets are mined only from initialization rollouts; held-out and
  out-of-distribution benchmarks are used for evaluation.}
  \label{tab:supp-benchmark-details}
\end{table*}

\subsection{Detailed Single-Operator Results}
\label{sec:supp-operator-results}

\begin{table*}[t]
  \centering
  \setlength{\tabcolsep}{1pt}
  \begin{tabular}{@{}>{\raggedright\arraybackslash}m{0.18\textwidth}
                  *{4}{>{\centering\arraybackslash}m{0.066\textwidth}}
                  *{2}{>{\centering\arraybackslash}m{0.095\textwidth}}
                  *{3}{>{\centering\arraybackslash}m{0.105\textwidth}}@{}}
    \toprule
    & \multicolumn{6}{c}{\textbf{Math}}
    & \multicolumn{3}{c}{\textbf{Code}} \\
    \cmidrule(lr){2-7} \cmidrule(lr){8-10}
    \textbf{Method}
    & \tablehead{\textbf{DAPO}\\\textbf{-hard}}
    & \tablehead{\textbf{AIME}\\\textbf{24}}
    & \tablehead{\textbf{AIME}\\\textbf{25}}
    & \tablehead{\textbf{AMC}\\\textbf{23}}
    & \tablehead{\textbf{Minerva}\\\textbf{Math}}
    & \tablehead{\textbf{Olympiad}\\\textbf{Bench}}
    & \tablehead{\textbf{APPS-}\\\textbf{train-hard}}
    & \tablehead{\textbf{APPS-}\\\textbf{test}}
    & \tablehead{\textbf{LiveCode}\\\textbf{Bench}} \\
    \midrule
    Revision-500 & 12.77 & 61.25 & \textbf{46.25} & 92.19 & 34.56 & 67.95 & 20.61 & 44.98 & \textbf{44.15} \\
    Cross-revision-500 & \textbf{13.50} & \textbf{62.92} & 45.42 & 91.56 & 34.93 & \textbf{69.29} & 21.12 & 44.39 & 43.12 \\
    Mutation-500 & 13.00 & 60.83 & \textbf{46.25} & 92.19 & \textbf{35.66} & 68.69 & \textbf{21.49} & \textbf{45.12} & 42.75 \\
    Crossover-500 & 13.09 & 61.25 & 45.00 & \textbf{92.81} & 34.93 & 69.14 & 21.42 & 44.64 & 43.23 \\
    Dual-reflection-500 & 13.27 & 61.25 & 45.83 & \textbf{92.81} & 34.56 & 68.10 & 20.66 & 44.83 & 43.96 \\
    \midrule
    Single-op-500 (avg.) & 13.13 & 61.50 & 45.75 & 92.31 & 34.93 & 68.63 & 21.06 & 44.79 & 43.44 \\
    \bottomrule
  \end{tabular}
  \caption{Detailed tiny-skill-bank single-operator results for Qwen3-4B.
  The average row is the arithmetic mean over the five independently evolved
  operator-specific banks. Code columns report mean unit-test pass rate (\%).}
  \label{tab:supp-single-operator-results}
\end{table*}

\subsection{Baseline Configuration}

\emph{No skill} removes skill retrieval from the solver context. \emph{Traj
RAG} retrieves trajectory-derived support but does not evolve it.
\emph{Reflexion} uses a reflection-style feedback update, and \emph{ExpeL-SI}
uses experience-summary-style skill induction. \emph{Init} uses the initialized
skill bank without evolution. \emph{Single-op} is the arithmetic mean across
the five independently evolved operator-specific banks, while \emph{Fusion}
uses \opfusion{} under the same overall bank-size budget. All methods are
evaluated with the same benchmark prompts and metric scripts within each
backbone block.

\subsection{Implementation Details}

Math skill banks are initialized from rollouts over roughly 17k DAPO problems,
and the approximately 4k failed problems form DAPO-hard for subsequent
evolution. Code skill banks are initialized from APPS training rollouts, and
evolution is applied to approximately 2.2k failed APPS training examples. Tiny
banks contain 500 initialized skills, while large banks use the larger curated
bank. Code uses execution-based pass-rate signals; math uses answer-level
correctness.

\paragraph{GRPO training configuration.}
Table~\ref{tab:supp-grpo-config} summarizes the optimization, batching, and
write-back settings used for the skill evolver.

\begin{table}[tbp]
  \centering
  \setlength{\tabcolsep}{2pt}
  \begin{tabular}{@{}lr@{}}
    \toprule
    \textbf{Setting} & \textbf{Value} \\
    \midrule
    Learning rate & $1.0\times10^{-6}$ \\
    Rollout batch & 32 \\
    Global update batch & 32 \\
    Update micro-batch / GPU & 2 \\
    Experience micro-batch / GPU & 8 \\
    Candidates per target & 4 \\
    Training epochs & 5 \\
    Explicit step cap & None \\
    Write-back cap & 2 / problem / round \\
    \bottomrule
  \end{tabular}
  \caption{GRPO training and write-back settings for the skill evolver.}
  \label{tab:supp-grpo-config}
\end{table}

Each target problem produces four candidate child skills, which form its GRPO
comparison group. Training runs for five epochs without a separate maximum-step
limit, so the epoch schedule determines the total number of updates. After
evaluation, at most two accepted child skills per problem are written back in
each evolution round. The global update batch matches the rollout batch, while
the update and experience phases use different per-GPU micro-batches.

\section{Complementary Skill Bank Fusion Details}
\label{sec:supp-fusion}

For the full-size fused banks, the target size is $N=22{,}133$ for math and
$N=7{,}644$ for code. We set the failure-like record ratio to $\rho=0.15$
and retain at most $B_q=2$ raw skill records for any problem identifier.

Each raw skill record $z$ has an operator source $o(z)$, problem identifier
$q(z)$, score $\mathrm{score}(z)$, correctness metadata, and optional textual
fields such as procedural steps, watchouts, and solution traces. We use
$\mathrm{Inc}(z)$ and $\mathrm{Corr}(z)$ for incorrect and correct metadata. A
record is reliable when $\neg\mathrm{Inc}(z)$ and either
$\mathrm{score}(z)\ge 0.5$ or $\mathrm{Corr}(z)$ holds. A record is
failure-like when $\mathrm{Inc}(z)$ holds or $\mathrm{score}(z)<0.5$.
Failure-like records are not used as positive procedures during retrieval, but
they may still yield useful \emph{Guard} or \emph{Validation} units.

\begin{appendixalgorithm}{Operator-Fused QID-Covered Skill Consolidation}
  1: & \textbf{Input:} operator skill sets $\{\mathcal{S}^{(o)}\}_{o\in\mathcal{O}}$, domain-specific target size $N$, failure ratio $\rho=0.15$, max skills per question $B_q=2$. \textbf{Output:} consolidated raw skill bank $\mathcal{S}^{\mathrm{mix}}$. \\
  2: & Initialize $\mathcal{S}^{\mathrm{mix}}\leftarrow\emptyset$ and soft operator quotas $Q_o\approx N/|\mathcal{O}|$. \\
  3: & Normalize all records; compute $q(z)$, reliability, failure-like flag, richness score, and source hash. \\
  4: & Deduplicate by source hash and group candidates by $q(z)$. \\
  5: & \textbf{Stage 1: QID coverage.} For each $q$ ordered by best candidate quality, add the best reliable and informative record, preferring under-quota operators. \\
  6: & \textbf{Stage 2: failure evidence.} While $|\mathcal{S}^{\mathrm{mix}}|<N$ and failure count is below $\rho N$, add the best remaining failure-like record with non-empty procedural evidence. \\
  7: & Require every added record to satisfy deduplication and $q(z)$ count $<B_q$. \\
  8: & \textbf{Stage 3: complementary reliable records.} Fill remaining slots with reliable records, preferring under-quota operators and under-covered problem identifiers. \\
  9: & \textbf{Stage 4: final fill.} Add the best remaining valid informative records. \\
  10: & \textbf{return} $\mathcal{S}^{\mathrm{mix}}$. \\
\end{appendixalgorithm}

For the 500-skill setting, an additional diversity step within each operator
pool prevents the final bank from concentrating on one semantic region. We use
$K=500$, reliable ratio $\alpha=0.8$ (and hence failure-like ratio $0.2$), and
$B_q=1$. Before diversity selection, the quality-ranked candidate pool is
capped at five times the corresponding selection quota. Diversity is measured
using each record's precomputed 384-dimensional memory vector.

\begin{appendixalgorithm}{Small-Bank Vector-Diverse Fusion}
  1: & \textbf{Input:} operator skill sets $\{\mathcal{S}^{(o)}\}_{o\in\mathcal{O}}$, size $K=500$, reliable ratio $\alpha=0.8$, per-question cap $B_q=1$, precomputed 384-dimensional vectors $v(z)$. \textbf{Output:} diversified bank $\mathcal{S}^{\mathrm{small}}$. \\
  2: & Set $\mathcal{S}^{\mathrm{small}}\leftarrow\emptyset$ and per-operator quota $Q_o=K/|\mathcal{O}|$. \\
  3: & \textbf{for each} operator $o\in\mathcal{O}$ \textbf{do} \\
  4: & \quad Split $\mathcal{S}^{(o)}$ into reliable candidates $R_o$ and failure-like candidates $F_o$. \\
  5: & \quad Keep at most one best candidate per problem identifier in each split and retain a quality pool of at most $5$ times the corresponding selection quota. \\
  6: & \quad Select about $\alpha Q_o$ reliable records by farthest-first traversal over normalized vectors $v(z)$. \\
  7: & \quad Select about $(1-\alpha)Q_o$ failure-like records with the same rule. \\
  8: & \quad Add selected records while enforcing global deduplication and one record per problem identifier. \\
  9: & \textbf{end for} \\
  10: & If $|\mathcal{S}^{\mathrm{small}}|<K$, fill remaining slots with globally diverse high-quality candidates. \\
  11: & \textbf{return} $\mathcal{S}^{\mathrm{small}}$. \\
\end{appendixalgorithm}

After fusion, selected raw skill records are proceduralized into typed units.
The retrieval bank filters positive and negative units differently.

\begin{appendixalgorithm}{Role-Consistent Filtering After Proceduralization}
  1: & \textbf{Input:} candidate procedural units $\mathcal{U}$ with source record $z(u)$ and role $r(u)$. \textbf{Output:} filtered retrieval bank $\mathcal{R}$. \\
  2: & Initialize $\mathcal{R}\leftarrow\emptyset$. \\
  3: & \textbf{for each} unit $u\in\mathcal{U}$ \textbf{do} \\
  4: & \quad \textbf{if} $r(u)\in\{\textsc{Guard},\textsc{Validation}\}$ \textbf{then} add $u$ to $\mathcal{R}$. \\
  5: & \quad \textbf{else if} $r(u)\in\{\textsc{Framing},\textsc{Strategy},\textsc{Execution}\}$ and $z(u)$ is reliable \textbf{then} add $u$ to $\mathcal{R}$. \\
  6: & \quad \textbf{end if} \\
  7: & \textbf{end for} \\
  8: & \textbf{return} $\mathcal{R}$. \\
\end{appendixalgorithm}

\newpage

\section{Procedural Skill Interface}
\label{sec:supp-psi}

The Procedural Skill Interface (PSI) is used at retrieval time to make evolved skills usable on unseen problems.
It introduces a shared procedural language between the skill bank and the target problem: evolved skills are converted into typed procedural units, while each target problem is converted into typed procedural keys for retrieval.  We use the paper-facing roles Framing, Strategy, Guard, Validation, and Execution, which correspond to the implementation types Modeling, Core, Trap, Verification, and Implementation, respectively; Execution is used only for code.

\begin{appendixalgorithm}{Procedural Skill Interface}
    1: & \textbf{Input:} evolved skill bank $\mathcal{S}$, target problem $x$, and domain $d$. \textbf{Output:} retrieval context $\mathcal{C}(x)$. \\
    2: & \textbf{Skill-side proceduralization.} For each skill $s \in \mathcal{S}$, generate procedural units $U_s \leftarrow \mathrm{SkillPrompt}(s,d)$. \\
    3: & Normalize each unit as (Role, Procedural Key, Trigger, Routine, Guard, Validation). \\
    4: & Remove source-specific constants, stories, row IDs, and answer values. \\
    5: & Keep Guard and Validation units; keep Framing, Strategy, and Execution units only when their source is reliable. \\
    6: & Encode each retained unit key with the locally hosted BGE-M3 encoder, $\ell_2$-normalize its vector, and add it to the role-specific index $\mathcal{I}_{r(u)}$. \\
    7: & \textbf{Query-side proceduralization.} Extract procedural keys $Q(x) \leftarrow \mathrm{QueryPrompt}(x,d)$. \\
    8: & Encode and $\ell_2$-normalize the query keys; consolidate query subquestions whose cosine similarity is at least $0.92$. \\
    9: & \textbf{for each} remaining query key $q \in Q(x)$ \textbf{do} \\
    10: & \quad Retrieve the top 10 role-compatible units from $\mathcal{I}_{r(q)}$, ranked by the dot product of normalized vectors (cosine similarity). \\
    11: & \textbf{end for} \\
    12: & Merge the retrieved candidates in descending similarity-score order. \\
    13: & Apply role caps (Framing: 1, Strategy: 1, Guard: 1, Validation: 0, Execution: 1) and retain at most five units overall. \\
    14: & Compose solver-facing Routines, Guards, and Validation cues, then \textbf{return} $\mathcal{C}(x)$. \\
\end{appendixalgorithm}

All compared skill banks use this same retrieval setup.  The top-10 search has
no additional relevance-score cutoff.  The $0.92$ threshold is used only to
consolidate query-side subquestions; it is not an embedding-based duplicate
threshold for retrieved skills.  The five-unit context limit is an outer cap,
so the role caps can produce fewer than five units.

\paragraph{Prompt templates.}
The prompts below show the domain-conditioned templates used to construct the PSI.  Math uses the roles Framing, Strategy, Guard, and Validation; code additionally uses Execution for implementation-sensitive details.

\newpage
\begin{promptbox}[\textbf{Prompt (Skill-Side Proceduralization)}]
SYSTEM:
You curate reusable procedural skills for math and code.

USER:
Convert one evolved skill into 0-3 units that help with
unseen same-domain problems, not the source problem itself.

GOAL:
- Preserve transferable procedures, constraints,
  invariants, failure modes, and checks.
- Remove source stories, constants, row IDs, exact
  answers, dataset names, long reasoning, and source code.

UNIT FIELDS:
- Role: Framing, Strategy, Guard, Validation, or Execution.
- Procedural Key: a "How do I ...?" retrieval question.
- Trigger: when the unit applies.
- Routine: reusable actions for the solver.
- Guard: a concrete mistake or false route to avoid.
- Validation: a concrete result or implementation check.

ROLE GUIDE:
- Framing: variables, states, constraints, or models.
- Strategy: method, recurrence, invariant, or reduction.
- Guard: false assumptions, shortcuts, or boundary loss.
- Validation: candidates, invariants, cases, or outputs.
- Execution: indexing, I/O, updates, or formatting.

RELIABILITY:
Failed or low-score sources normally yield only Guard or
Validation units. Use positive roles only for reliable evidence.

INPUT:
DOMAIN: {domain}
SOURCE_PROBLEM: {problem_text}
SOURCE_SCORE: {score}
SOURCE_IS_CORRECT: {is_correct}
EVOLVED_STEPS: {steps}
EVOLVED_WATCHOUTS: {watchouts}
OPTIONAL_SOLUTION_OR_CODE: {solution_or_code}

OUTPUT:
JSON only: 0-3 units with all six fields above.
\end{promptbox}

\newpage
\begin{promptbox}[\textbf{Prompt (Query-Side Proceduralization)}]
SYSTEM:
You extract procedural retrieval keys for unseen math
or coding problems.

USER:
From only the target statement, produce 2-4 keys for
retrieving useful procedural units.

RULES:
- Do not solve the problem or compute the final answer.
- Do not overfit to samples, constants, labels, or stories.
- Prefer a few high-signal keys over broad coverage.
- Be more general than the exact problem but more
  specific than "use DP" or "check cases".

KEY FIELDS:
- Role: Framing, Strategy, Guard, Validation, or Execution.
- Procedural Key: a stable "How do I ...?" question.

ROLE GUIDE:
- Framing: model, variables, states, or constraints.
- Strategy: a transferable method without over-commitment.
- Guard: likely false assumptions or traps.
- Validation: tests for the answer or algorithm.
- Execution: indexing, I/O, or state-update details.

INPUT:
DOMAIN: {domain}
PROBLEM_STATEMENT: {problem_statement}

OUTPUT:
JSON only: 2-4 keys with Role and Procedural Key.
\end{promptbox}
\newpage

\section{Evolution Operator Details}
\label{sec:supp-operator-details}

The following algorithm blocks summarize how each operator selects support
skills and renders its child skill.


\paragraph{Revision.}
\begin{appendixalgorithm}{Same-Problem Conservative Revision}
    1: & \textbf{Input:} target skill $s$, skill bank $\mathcal{B}$. \textbf{Output:} revised child skill $s'$. \\
    2: & $C \leftarrow \mathrm{SearchByQID}(\mathcal{B}, s.qid)$. \\
    3: & Remove target skill $s$ and skills marked inactive from $C$. \\
    4: & $C_{\mathrm{correct}} \leftarrow \{c \in C : c.\mathrm{is\_correct}\}$. \\
    5: & $C_{\mathrm{orig}} \leftarrow \{c \in C : c.\mathrm{is\_original}\}$. \\
    6: & \textbf{if} $C_{\mathrm{correct}} \neq \emptyset$ \textbf{then} \\
    7: & \quad $b \leftarrow \arg\max_{c \in C_{\mathrm{correct}}}(c.\mathrm{score},c.\mathrm{version})$. \\
    8: & \textbf{else if} $C_{\mathrm{orig}} \neq \emptyset$ \textbf{then} \\
    9: & \quad $b \leftarrow \arg\max_{c \in C_{\mathrm{orig}}}(c.\mathrm{score},c.\mathrm{version})$. \\
    10: & \textbf{else} $b \leftarrow \arg\max_{c \in C}(c.\mathrm{score},c.\mathrm{version})$. \\
    11: & $F \leftarrow \{c \in C : \neg c.\mathrm{is\_correct} \land c \neq b\}$. \\
    12: & $f \leftarrow \arg\max_{c \in F}(c.\mathrm{score},c.\mathrm{version})$. \\
    13: & $P \leftarrow \mathrm{RenderRevisionPrompt}(s,b,f)$. \\
    14: & $s' \leftarrow \mathrm{LLM}(P)$; \textbf{return} $s'$. \\
\end{appendixalgorithm}

\paragraph{Cross-revision.}
\begin{appendixalgorithm}{Cross-Problem Transfer-Based Conservative Revision}
    1: & \textbf{Input:} target skill $s$, skill bank $\mathcal{B}$. \textbf{Output:} cross-revised child skill $s'$. \\
    2: & $C \leftarrow \mathrm{SemanticSearch}(\mathcal{B}, s.\mathrm{problem})$. \\
    3: & Remove same-qid, inactive, and duplicate candidates from $C$. \\
    4: & Partition $C$ into tagged/fallback positive and negative pools. \\
    5: & $r \leftarrow \mathrm{PickTop}(C_{\mathrm{tagged}}^{+}, C_{\mathrm{fallback}}^{+})$. \\
    6: & $t \leftarrow \mathrm{PickTop}(\text{negative pools};\mathrm{exclude}=r)$. \\
    7: & \textbf{if} $t=\emptyset$ \textbf{then} choose best remaining $c\in C\setminus\{r\}$. \\
    8: & $P \leftarrow \mathrm{RenderCrossRevisionPrompt}(s,r,t)$. \\
    9: & $s' \leftarrow \mathrm{LLM}(P)$; \textbf{return} $s'$. \\
\end{appendixalgorithm}

\paragraph{Mutation.}
\begin{appendixalgorithm}{Mutation Operator}
    1: & \textbf{Input:} target skill $s$, skill bank $\mathcal{B}$. \textbf{Output:} mutated child skill $s'$. \\
    2: & $C_{\mathrm{same}} \leftarrow \mathrm{SearchByQID}(\mathcal{B},s.qid)$. \\
    3: & $C_{\mathrm{sem}} \leftarrow \mathrm{SemanticSearch}(\mathcal{B},s.\mathrm{problem})$. \\
    4: & Remove $s$, inactive candidates, same-qid semantic candidates, and duplicates. \\
    5: & Partition $C_{\mathrm{sem}}$ into correct and incorrect semantic pools. \\
    6: & $A \leftarrow \{c \in C_{\mathrm{same}} : \neg c.\mathrm{is\_correct}\}$. \\
    7: & $r \leftarrow \mathrm{PickTop}(A, C_{\mathrm{sem}}^{-})$. \\
    8: & $d \leftarrow \mathrm{PickTop}(C_{\mathrm{sem}}^{+};\mathrm{exclude}=r)$. \\
    9: & $P \leftarrow \mathrm{RenderMutationPrompt}(s,r,d)$. \\
    10: & $s' \leftarrow \mathrm{LLM}(P)$; \textbf{return} $s'$. \\
\end{appendixalgorithm}

\paragraph{Dual-reflection.}
\begin{appendixalgorithm}{Guard-Oriented Reflective Enhancement}
    1: & \textbf{Input:} target skill $s$, skill bank $\mathcal{B}$, fallback flag $\rho$. \textbf{Output:} reflected child skill $s'$. \\
    2: & $C_{\mathrm{same}} \leftarrow \mathrm{SearchByQID}(\mathcal{B},s.qid)$. \\
    3: & $C_{\mathrm{sem}} \leftarrow \mathrm{SemanticSearch}(\mathcal{B},s.\mathrm{problem})$. \\
    4: & Remove $s$, inactive candidates, same-qid semantic candidates, and duplicates. \\
    5: & Partition semantic candidates into correct and incorrect pools. \\
    6: & $u \leftarrow \mathrm{PickTop}(C_{\mathrm{sem}}^{+})$. \\
    7: & $A \leftarrow \{c \in C_{\mathrm{same}} : \neg c.\mathrm{is\_correct} \land c \neq u\}$. \\
    8: & $t \leftarrow \mathrm{PickTop}(A,C_{\mathrm{sem}}^{-};\mathrm{exclude}=u)$. \\
    9: & \textbf{if} $u=\emptyset$ and $t=\emptyset$ and $\rho=\mathrm{true}$ \textbf{then return} $\mathrm{Mutation}(s,\mathcal{B})$. \\
    10: & $P \leftarrow \mathrm{RenderDualReflectionPrompt}(s,u,t)$. \\
    11: & $s' \leftarrow \mathrm{LLM}(P)$; \textbf{return} $s'$. \\
\end{appendixalgorithm}

\paragraph{Crossover.}
\begin{appendixalgorithm}{Same-Problem Two-Parent Crossover}
    1: & \textbf{Input:} target skill $s$, skill bank $\mathcal{B}$, parent limit $L$. \textbf{Output:} crossover child skill $s'$. \\
    2: & $C \leftarrow \mathrm{SearchByQID}(\mathcal{B},s.qid,L)$. \\
    3: & Remove $s$, inactive/original/incorrect skills, empty plans, and duplicate plan hashes. \\
    4: & \textbf{if} $|C|<2$ \textbf{then} \\
    5: & \quad \textbf{if} $|C|=1$ \textbf{then return} $\mathrm{Revision}(s,\mathcal{B})$. \\
    6: & \quad \textbf{else return} $\mathrm{DualReflection}(s,\mathcal{B})$. \\
    7: & Sort $C$ by $(\mathrm{score},\mathrm{version})$ in descending order. \\
    8: & $p_b \leftarrow C[1]$. \\
    9: & Define $q(c)$ by step difference, distance to $0.25$, and negative score. \\
    10: & $p_c \leftarrow \arg\min_{c \in C\setminus\{p_b\}} q(c)$. \\
    11: & $P \leftarrow \mathrm{RenderCrossoverPrompt}(s,p_b,p_c)$. \\
    12: & $s' \leftarrow \mathrm{LLM}(P)$; \textbf{return} $s'$. \\
\end{appendixalgorithm}

\subsection{Compact Operator Prompt Specifications}

For legibility, we factor the repeated output, schema, and content constraints
into one shared contract, followed by one operator-specific block per
operator. The deployed prompt is the concatenation of the shared contract and
the selected operator block. This presentation removes repeated boilerplate
and decorative separators while preserving the operator rules and
implementation-specific variables in Jinja form. The source package also
retains the unabridged templates.


The runtime prompt is formed by prepending the shared contract below to the
selected operator-specific block. This factorization removes repeated
boilerplate and decorative separators while retaining the task, constraints,
input variables, and output contract.

\newpage
\begin{promptbox}[\textbf{Shared Prompt Contract}]
ROLE:
You are a Strategy Skill Editor.

GOAL:
Do not solve the target problem. Produce a reusable,
concrete, information-dense method-level plan that a
separate solver can execute. Avoid long derivations,
low-level code narration, and line-by-line instructions.

OUTPUT:
Return exactly:
<think> short operator-focused reasoning </think>
followed by one valid JSON object and nothing else.

JSON RULES:
- No markdown, code fences, extra text, or trailing commas.
- Use double quotes for every key and string value.
- Put all actionable content in the JSON object.

SCHEMA:
{
  "steps": [{"i": 1, "text": "string"}],
  "watchouts": ["string"]
}

CONTENT RULES:
- Prefer 4-5 steps and 2-3 watchouts.
- Use at most 6 steps and at most 3 watchouts.
- Each step states one concrete method-level action.
- Watchouts are distinct, realistic guardrails.
- Protect correctness, contract, state, boundaries,
  implementation stability, and complexity as applicable.
- Keep the result reusable across related problems.
\end{promptbox}

\newpage
\begin{promptbox}[\textbf{Operator Block (Revision)}]
TASK:
Conservatively repair the current weak strategy.

THINK:
Identify the smallest useful repair. Do not solve the
problem or replace a usable backbone.

REVISION RULES:
- Stay in the same algorithm family when possible.
- Use BACKBONE SUPPORT to stabilize the main route.
- Use FAILURE SUPPORT to patch one weak point.
- Prefer a local bridge, ordering, guard, or validation fix.
- Replace the global route only if it is unusable.

APPS PRIORITY:
- Treat revision as watchouts-first conservative repair.
- Change steps less than watchouts when the route works.
- Add a contract, complexity, boundary, or state guard.

INPUT FIELDS (Jinja):
problem: {{ problem }}
current_weak_plan: {{ current_plan }}
current_watchouts: {{ target_failed_watchouts }}
  [if present]
backbone_support: {{ backbone_support }}
backbone_watchouts: {{ backbone_support_watchouts }}
  [if present]
failure_support: {{ failure_support }}
failure_watchouts: {{ failure_support_watchouts }}
  [if present]

RETURN:
One revised strategy object under the shared contract.
\end{promptbox}

\newpage
\begin{promptbox}[\textbf{Operator Block (Cross-revision)}]
TASK:
Improve the target strategy using cross-problem support.

THINK:
Choose one transferable motif and one realistic guard.
Do not solve the target problem.

CROSS-REVISION RULES:
- Treat this as conservative cross-motif transfer.
- Never transplant a whole foreign route by default.
- Keep the target backbone primary.
- Borrow at most one compatible donor-local motif:
  a maintenance trick, check pattern, complexity trick,
  or contract-oriented watchout.
- ROUTE DONOR may supply one transferable motif.
- TRAP DONOR may supply one watchout or small repair.
- Reject donors that conflict with the target contract.
- Do not output a stitched list of donor steps.

APPS PRIORITY:
Transfer only when it improves contract, indexing, state,
complexity, or output consistency.

INPUT FIELDS (Jinja):
problem: {{ problem }}
current_weak_plan: {{ current_plan }}
current_watchouts: {{ target_failed_watchouts }}
  [if present]
route_donor: {{ route_donor }} [if present]
route_donor_watchouts: {{ route_donor_watchouts }}
  [if present]
trap_donor: {{ trap_donor }} [if present]
trap_donor_watchouts: {{ trap_donor_watchouts }}
  [if present]

RETURN:
One cross-revised strategy object under the shared contract.
\end{promptbox}

\newpage
\begin{promptbox}[\textbf{Operator Block (Mutation)}]
TASK:
Replace the weak strategy with a stronger alternative.

THINK:
Identify one justified backbone change in algorithm family,
decomposition, or complexity tier. Do not solve the problem.

MUTATION RULES:
- Treat CURRENT WEAK PLAN as the baseline to replace.
- Use ROUTE TO AVOID as negative evidence.
- Use DIVERGENT SEED for one stronger backbone idea.
- Rebuild steps and watchouts as one coherent package.
- Do not merely paraphrase the current plan.
- Reject a seed that repeats the route to avoid.
- Do not generate a random unrelated route.

APPS PRIORITY:
Prefer changes that fix the algorithm family, complexity
tier, decision/reconstruction contract, or state design.
A new route must include its new failure modes.

INPUT FIELDS (Jinja):
problem: {{ problem }}
current_weak_plan: {{ current_plan }}
current_watchouts: {{ target_failed_watchouts }}
  [if present]
route_to_avoid: {{ route_to_avoid }}
route_to_avoid_watchouts: {{ route_to_avoid_watchouts }}
  [if present]
divergent_seed: {{ divergent_seed }}
divergent_seed_watchouts: {{ divergent_seed_watchouts }}
  [if present]

RETURN:
One alternative strategy object under the shared contract.
\end{promptbox}

\newpage
\begin{promptbox}[\textbf{Operator Block (Dual-reflection)}]
TASK:
Keep one coherent route while strengthening its guardrails.

THINK:
Identify the strongest realistic failure mode and the
smallest guard-heavy repair. Do not solve the problem.

DUAL-REFLECTION RULES:
- Keep plausible steps mostly stable and edits local.
- Rewrite or strengthen watchouts as the main operation.
- Provide three concrete watchouts.
- TRAP SUPPORT is direct evidence for failure guards.
- MECHANISM SUPPORT may add one clean forward mechanism.
- Trap support alone is sufficient when it is diagnostic.
- Do not invent a second competing route.

WATCHOUT PRIORITY:
Cover invariant/state-reset safety, boundary/indexing,
interface/contract safety, and complexity/contamination.

INPUT FIELDS (Jinja):
problem: {{ problem }}
target_failed_strategy: {{ target_failed_strategy }}
target_watchouts: {{ target_failed_watchouts }}
  [if present]
mechanism_support: {{ mechanism_support }} [if present]
mechanism_watchouts: {{ mechanism_support_watchouts }}
  [if present]
trap_support: {{ trap_support }} [if present]
trap_watchouts: {{ trap_support_watchouts }}
  [if present]

RETURN:
One reflected strategy object under the shared contract.
\end{promptbox}

\newpage
\begin{promptbox}[\textbf{Operator Block (Crossover)}]
TASK:
Combine one stable backbone with one small improvement.

THINK:
Identify the backbone and one compatible motif to borrow.
Do not solve the target problem.

CROSSOVER RULES:
- Build around BACKBONE PARENT as the main route.
- Use CONTRAST PARENT for at most one local idiom,
  maintenance trick, or guard motif.
- Never replace the backbone through free-form union.
- Do not splice incompatible states, data structures,
  invariants, or reconstruction contracts.
- If compatibility is unclear, keep the backbone stable.
- Regenerate watchouts for the merged child.

APPS PRIORITY:
Prefer contributions improving contract stability,
boundary/indexing safety, state maintenance, or complexity
rather than novelty alone.

INPUT FIELDS (Jinja):
problem: {{ problem }}
backbone_parent: {{ backbone_parent }}
backbone_watchouts: {{ backbone_parent_watchouts }}
  [if present]
contrast_parent: {{ contrast_parent }}
contrast_watchouts: {{ contrast_parent_watchouts }}
  [if present]

RETURN:
One crossover strategy object under the shared contract.
\end{promptbox}
\newpage



\end{document}